\documentclass[letterpaper]{article} 
\usepackage[preprint]{aaai2027}  
\usepackage[hyphens]{url}  
\usepackage{graphicx} 
\usepackage{natbib}  
\usepackage{caption} 
\usepackage{algorithm}
\usepackage{algorithmic}

\usepackage{newfloat}
\usepackage{listings}
\DeclareCaptionStyle{ruled}{labelfont=normalfont,labelsep=colon,strut=off} 
\floatstyle{ruled}
\newfloat{listing}{tb}{lst}{}
\floatname{listing}{Listing}

\usepackage{booktabs}

\usepackage{amsfonts}       
\usepackage{nicefrac}       
\usepackage{microtype}      
\usepackage[table]{xcolor}
\usepackage{multirow}
\usepackage{array}
\usepackage{makecell}
\usepackage{subcaption}
\usepackage{amsmath}
\usepackage{bbding}
\usepackage{tabularx}
\usepackage{xparse}
\usepackage{enumitem}

\definecolor{oursblue}{HTML}{F1F4F9}
\definecolor{improveblue}{HTML}{3282B8}
\definecolor{improvegreen}{HTML}{2A7F62}
\definecolor{groupgray}{HTML}{F2F2F2}

\newcommand{\thinvrule}{%
    {\color{black!60}\vrule width 0.25pt}%
}

\newcommand{\methodrule}{%
    \noalign{\vskip 1.1pt}%
    \specialrule{0.30pt}{0pt}{0pt}%
    \noalign{\vskip 1.1pt}%
}

\newcommand{\improvement}[1]{\textnormal{ (#1)}}
\newcommand{\gain}[1]{\textnormal{ (#1)}}

\newcommand{\rowrule}{%
    \specialrule{0.20pt}{0pt}{0pt}%
}

\usepackage{placeins}
\newcommand{\epdms}[2]{%
    \makebox[2.35em][r]{#1}%
    \makebox[2.60em][l]{\improvement{#2}}%
}
\title{PAD: Policy-Aware Data Valuation for Autonomous Driving}
\author{
    Chengkai Xu\textsuperscript{1},
    Yicheng Guo\textsuperscript{1},
    Jiaqi Liu\textsuperscript{2},
    Yiming Cui\textsuperscript{1},\\
    Shiyu Fang\textsuperscript{1},
    Peng Hang\textsuperscript{1},
    Hao Zhang\textsuperscript{3},
    Jian Sun\textsuperscript{1}\corresponding
}
\affiliations{
    \textsuperscript{1}College of Transportation, Tongji University\\
    \textsuperscript{2}Department of Computer Science, UNC Chapel Hill\\
    \textsuperscript{3}Department of Control Science and Engineering, Tongji University
}

\begin{document}

\maketitle

\begin{abstract}
\label{abstract}
End-to-end autonomous driving increasingly depends on scaling training demonstrations, yet data growth does not guarantee proportional gains. 
However, additional data do not guarantee better policies: a demonstration that benefits one planner may be redundant or harmful to another.
Existing curation methods typically infer data value from intrinsic sample properties or training-dependent proxies that do not measure downstream policy utility.
To address this, we introduce \textbf{Policy-Aware Data Valuation (PAD)}, a framework that assigns each candidate demonstration a utility specific to the target driving policy.
PAD combines scene-level calibration feedback with projected policy gradients and a regularized approximation of local training geometry, producing scalable influence-based values for the policy.
A controlled group-removal study establishes the directional fidelity of these values, while cross-planner analysis shows that the same candidate pool induces markedly different rankings across planner families.
On DiffusionDrive, PAD improves EPDMS over real-only training by $3.5$ points on NavTest and $5.0$ points on NavHard.
Policy-specific curation consistently outperforms real-only and full-pool training across diffusion-, regression-, and flow-matching-based planners while substantially reducing scene-level regressions.
These results establish policy-aware data valuation as a basis for replacing indiscriminate data accumulation with targeted policy improvement, enabling more efficient and reliable scaling of autonomous-driving systems.
\end{abstract}

\begin{links}
    \link{Code}{https://github.com/PerfectXu88/PAD}
    \link{Project Page}{https://perfectxu88.github.io/PAD/}
\end{links}


\section{Introduction}
\label{introduction}

End-to-end autonomous driving has advanced rapidly through increasingly capable policy architectures~\cite{zheng2024genad,chen2024end}, yet attainable performance remains constrained by the quality and composition of their training data~\cite{feng2026breaking}. 
The success of large-scale vision and language models has consequently motivated the continued expansion of real and synthetic driving corpora~\cite{jiang2025survey}. 
Data volume, however, does not translate predictably into policy improvement. 
Additional demonstrations may be redundant, inconsistent with the learning objective, or harmful because of distribution and supervision errors~\cite{shumailov2024ai,jiao2026learning, atakishiyev2025safety}. 
Such data can produce diminishing returns and scenario-specific regressions even when aggregate performance improves. 
The central data problem is therefore not simply how much to collect, but which demonstrations improve a given driving policy~\cite{dimlioglu2026scaling,xu2026survey}.

Addressing this problem requires a policy-dependent definition of data value. 
Research on data attribution and model-aware filtering increasingly evaluates training samples through their effects on learned models~\cite{yu2024mates, guo2026evaluation}. 
This perspective reveals that a demonstration has no policy-independent utility. 
The same sample may reinforce useful behavior in one policy while exerting negligible or adverse influence in another~\cite{choe2026your,hammoudeh2024training}. 
We accordingly define \emph{policy-aware data valuation} as estimating the downstream utility of an individual demonstration with respect to the target policy that learns from it.

\begin{figure*}
    \centering
    \includegraphics[width=\linewidth]{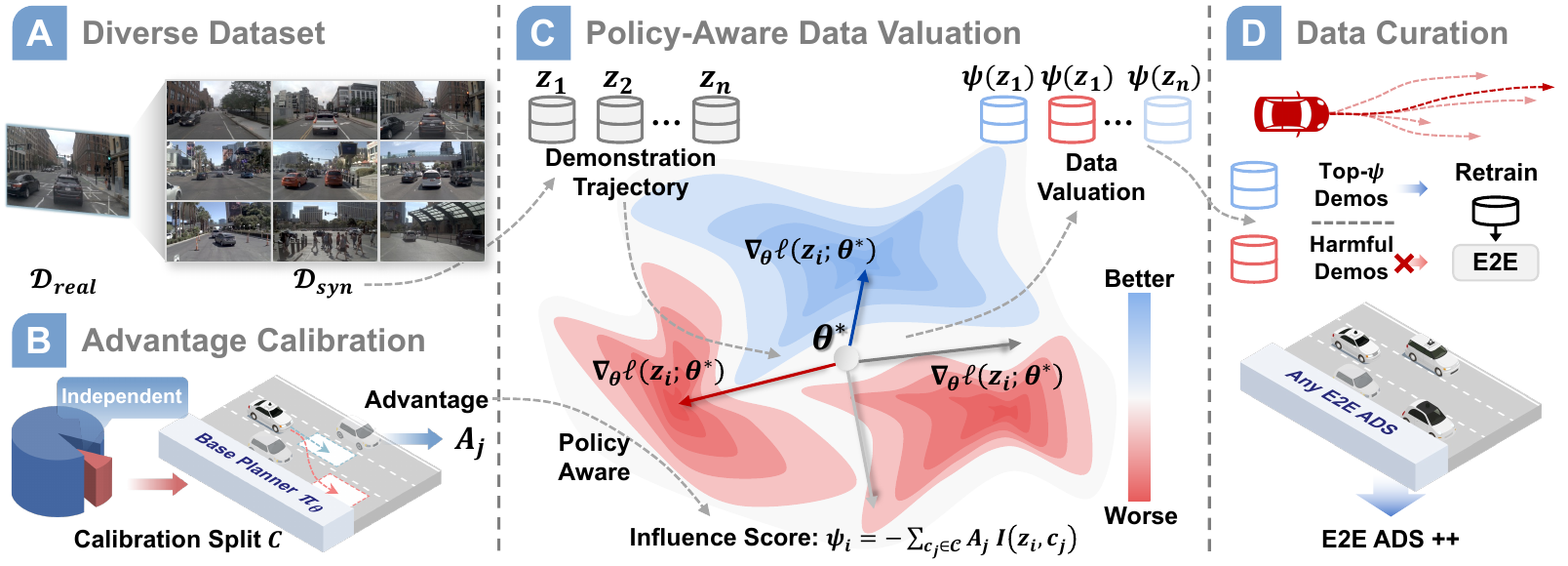}
    \caption{Overview of PAD, which combines scene-level advantages from an independent calibration split with projected planning gradients and local training geometry to estimate the planner-relative influence of each demonstration, enabling budgeted selection of beneficial data for retraining.}
    \label{fig:framework_overview}
\end{figure*}

Existing driving-data construction and curation methods capture only part of this policy dependence. 
Dataset-generation methods expand scale and coverage, usually treating complex or challenging scenarios as inherently valuable~\cite{rowe2025scenario, xiao2025robotron}. 
Data-intrinsic heuristics similarly prioritize rarity, diversity, or difficulty without measuring downstream learning effects~\cite{jiao2026learning}. 
However, scenario complexity is not equivalent to training utility and cannot reflect whether a demonstration induces a beneficial update for a particular policy. 
Training-dependent signals such as loss, uncertainty, and gradient salience reflect the current policy but need not correspond to downstream driving performance~\cite{luo2026unleashing,wang2024greats}. 
Dataset- or cluster-level filtering can further conceal substantial variation among demonstrations within the same group~\cite{calian2026datarater,chhabra2025outlier}. 
The unresolved challenge is a scalable, per-demonstration estimator that connects a candidate's effect on policy optimization to its downstream utility.

To address this challenge, \textbf{Policy-Aware Data Valuation (PAD)} is introduced. 
As illustrated in Fig.~\ref{fig:framework_overview}, PAD evaluates a target policy on an independent calibration set and uses the resulting scene-level outcomes to weight gradients of its native training objective. 
Projected gradients and a regularized approximation of local training geometry then yield a policy-specific value for each candidate demonstration, which supports budgeted selection for retraining. 
Controlled removal experiments show that these values predict realized training effects, while cross-policy analysis reveals distinct valuations of the same candidate pool. 
On DiffusionDrive, PAD improves EPDMS over real-only training by $3.5$ points on NavTest and $5.0$ points on NavHard, with consistent gains across multiple policy families and fewer scene-level regressions.

Our contributions are summarized as follows:
\begin{itemize}
    \item \textbf{Policy-aware data valuation.} We formulate the value of each driving demonstration relative to the target policy and its downstream utility, rather than as an intrinsic property of the sample.
    \item \textbf{Scalable per-demonstration estimation.} We develop an influence-based estimator that integrates scene-level policy outcomes, projected training gradients, and regularized local geometry within a unified interface across policy families.
    \item \textbf{Comprehensive empirical validation.} We show that PAD predicts realized training effects, produces distinct valuations across policies, improves downstream driving performance, and reduces scene-level regression exposure.
\end{itemize}

\section{Related Work}
\label{related_work}

\subsection{Driving Data Construction and Curation}

End-to-end driving systems increasingly rely on large and heterogeneous training corpora~\cite{zhou2026opendrivevla,kirby2026driving, xu2025knowledge}. 
Real-world fleet collection provides authentic driving behavior but remains costly, geographically constrained, and inherently sparse in safety-critical events~\cite{liu2025tad,wang2024drivingdojo}. 
Simulation and synthetic-data generation complement this process by expanding scenario coverage and enabling targeted construction of rare, interactive, or challenging situations~\cite{jiang2024scenediffuser,rowe2025scenario,kim2025synad}. 
These approaches have broadened the demonstrations available to driving models, but their primary objective is to construct the candidate data pool rather than determine how an individual demonstration will affect the policy trained on it.

As driving datasets grow, data curation methods seek to prioritize demonstrations according to intrinsic properties such as rarity, coverage or difficulty through sample selection, clustering, or reweighting~\cite{lu2026activead,embacher2026searchad}. 
Most of these methods implicitly regard complex or challenging scenarios as especially valuable~\cite{xiao2025robotron}. 
Although such criteria are useful for improving dataset composition, scenario complexity is not equivalent to training utility~\cite{xu2026tell}. 
A difficult demonstration may be redundant or harmful under imperfect supervision, while an ordinary demonstration may correct a consequential policy deficiency~\cite{ge2025unraveling}.
Data-intrinsic criteria therefore characterize the demonstration itself but cannot establish its downstream contribution to a target policy.

\subsection{Training Data Valuation}

Learner-aware curation methods address this limitation by evaluating samples through training-dependent signals, including loss, uncertainty, and gradient statistics~\cite{wang2024greats,xia2024less}. 
These signals reflect the state of the model, but they typically quantify difficulty, confidence, or optimization salience rather than the effect of a sample on the downstream objective of interest~\cite{mlodozeniec2026distributional}. 
Training data attribution provides a more direct counterfactual formulation by asking how perturbing or removing an individual example changes a target model behavior~\cite{ye2025towards}. 
Though these methods offer an empirical estimate of this effect, they can be prohibitively expensive at scale, especially in autonomous driving systems. 
Instead, influence functions and scalable gradient-based approximations estimate the response of a target quantity to infinitesimal changes in sample weight without retraining for every candidate, thereby defining data value relative to both a learner and an evaluation objective~\cite{tan2026understanding, agia2025cupid}.

Training data attribution has supported dataset debugging, behavior tracing, and data selection in supervised learning and large models~\cite{dass2025datamil,bae2024training}, and has recently been applied to robot demonstrations~\cite{agia2025cupid}. 
Per-demonstration valuation remains underexplored in autonomous driving, where high-dimensional sensory policies complicate attribution and supervised training losses do not directly represent downstream driving utility. Existing driving-data methods consequently rely largely on proxy scores or coarse-grained groups, leaving the planning effect of individual demonstrations unresolved~\cite{ruan2026closed}. 
In this paper, the proposed PAD addresses this gap by combining independent scene-level policy outcomes with scalable projected influence estimation to assign policy-specific values to individual driving demonstrations.

\section{Methodology}
\label{Methodology}

\subsection{Problem Formulation}
\label{sec:problem_formulation}

Let $\pi_{\theta}$ denote a target driving policy and $z=(o,\tau^{\star})$ a demonstration comprising an observation history $o$ and a reference trajectory $\tau^{\star}$. Given a real training set $\mathcal{D}_{\mathrm{real}}$, a candidate pool $\mathcal{D}_{\mathrm{pool}}=\{z_i\}_{i=1}^{N}$, and a budget $B$, define $\Omega_B=\{\mathcal{S}\subseteq\mathcal{D}_{\mathrm{pool}}:|\mathcal{S}|=B\}$ and $\theta_{\mathcal{S}}=\operatorname{Fit}(\mathcal{D}_{\mathrm{real}}\cup\mathcal{S})$. The optimal subset satisfies
\begin{equation}
    \mathcal{S}^{\star}_B
    =\underset{\mathcal{S}\in\Omega_B}{\arg\max}\;
    \mathcal{J}(\pi_{\theta_{\mathcal{S}}}),
    \label{eq:subset_selection_objective}
\end{equation}
where $\mathcal{J}$ denotes the expected downstream utility of the trained policy. Exact optimization requires retraining and evaluating a combinatorial number of subsets and is therefore intractable at driving-dataset scale.

PAD approximates this objective through local attribution around the base parameters $\hat{\theta}=\operatorname{Fit}(\mathcal{D}_{\mathrm{real}})$ and policy $\hat{\pi}=\pi_{\hat{\theta}}$. As shown in Fig.~\ref{fig:framework_overview}, PAD combines per-demonstration training gradients with scene-level outcomes from an independent calibration set $\mathcal{C}=\{z_j^{\mathrm{cal}}\}_{j=1}^{M}$. This construction assigns each candidate a policy-specific value $\psi_i$, from which the budgeted subset is selected.

\subsection{Local Influence Formulation}
\label{sec:gradient_representation}

We partition the policy parameters as $\theta=(\theta_{\mathrm{up}},\beta)$, where $\beta\in\mathbb{R}^{P}$ contains the trajectory-producing parameters used for attribution and $\theta_{\mathrm{up}}$ contains the upstream parameters. For a demonstration $z$, define its expected supervised trajectory loss and the gradient of candidate $z_i$ at the base checkpoint as
\begin{equation}
    \bar{\ell}(z;\beta)
    :=\mathbb{E}_{\xi}\!\left[
    \ell_{\mathrm{pl}}(z;\hat{\theta}_{\mathrm{up}},\beta,\xi)
    \right],
    \qquad
    q_i:=\nabla_{\beta}\bar{\ell}(z_i;\hat{\beta}),
    \label{eq:planning_gradient}
\end{equation}
where $\xi$ collects stochastic variables in the native training objective and $\hat{\theta}=(\hat{\theta}_{\mathrm{up}},\hat{\beta})$. With $\hat{\theta}_{\mathrm{up}}$ fixed, the local real-data objective is
\begin{equation}
    L_{\mathrm{real}}(\beta)
    :=\frac{1}{|\mathcal{D}_{\mathrm{real}}|}
    \sum_{z\in\mathcal{D}_{\mathrm{real}}}\bar{\ell}(z;\beta).
    \label{eq:local_real_objective}
\end{equation}
The upstream network is frozen only during attribution. Final curated retraining follows the original full-model training recipe.

For a candidate $z_i\in\mathcal{D}_{\mathrm{pool}}$, consider the add-one-in perturbation
\begin{equation}
    \hat{\beta}_{\eta,i}
    :=\underset{\beta}{\arg\min}\;
    \big[L_{\mathrm{real}}(\beta)+\eta\bar{\ell}(z_i;\beta)\big],
    \label{eq:candidate_perturbation}
\end{equation}
where $\eta$ is an infinitesimal candidate weight and $\hat{\beta}_{0,i}=\hat{\beta}$. Let $H:=\nabla_{\beta}^{2}L_{\mathrm{real}}(\hat{\beta})$. Under local stationarity and nonsingularity of $H$~\cite{koh2017understanding}, the first-order condition is
\begin{equation}
    \nabla_{\beta}L_{\mathrm{real}}(\hat{\beta}_{\eta,i})
    +\eta\nabla_{\beta}\bar{\ell}(z_i;\hat{\beta}_{\eta,i})=0.
    \label{eq:perturbed_stationarity}
\end{equation}
Differentiating Eq.~\ref{eq:perturbed_stationarity} at $\eta=0$ yields
\begin{equation}
    H\left.\frac{d\hat{\beta}_{\eta,i}}{d\eta}\right|_{\eta=0}+q_i=0,
    \qquad
    \left.\frac{d\hat{\beta}_{\eta,i}}{d\eta}\right|_{\eta=0}
    =-H^{-1}q_i.
    \label{eq:parameter_response}
\end{equation}

For calibration scene $z_j^{\mathrm{cal}}$, let
\begin{equation}
    f_j(\beta)
    :=\mathbb{E}_{\xi}\!\left[
    \ell_{\mathrm{pl}}(z_j^{\mathrm{cal}};
    \hat{\theta}_{\mathrm{up}},\beta,\xi)
    \right],
    \qquad
    q_j^{\mathrm{cal}}:=\nabla_{\beta}f_j(\hat{\beta}).
    \label{eq:calibration_loss}
\end{equation}
These are supervised-loss gradients evaluated on reference trajectories, rather than policy-score gradients from on-policy rollouts.
Applying the chain rule to Eq.~\ref{eq:parameter_response}, the influence of candidate $i$ on calibration loss $j$ is
\begin{equation}
    I_{ij}
    :=\left.\frac{d f_j(\hat{\beta}_{\eta,i})}{d\eta}\right|_{\eta=0}
    =-(q_j^{\mathrm{cal}})^{\top}H^{-1}q_i,
    \qquad
    g_{ij}:=-I_{ij}.
    \label{eq:pairwise_influence}
\end{equation}
Thus, $g_{ij}>0$ indicates that infinitesimally upweighting $z_i$ is locally predicted to reduce the supervised trajectory loss on calibration scene $j$.

\subsection{Policy-Aware Calibration}
\label{sec:calibration_direction}

Pairwise helpfulness measures how a candidate affects supervised loss on a calibration scene, but does not determine whether the corresponding behavior is valuable to the target policy. We therefore evaluate $\hat{\pi}$ on each $z_j^{\mathrm{cal}}\in\mathcal{C}$ and standardize its scene-level outcome:
\begin{equation}
    s_j=\operatorname{EPDMS}(\hat{\pi};z_j^{\mathrm{cal}}),
    \qquad
    A_j=\frac{s_j-\mu_{\mathcal{C}}}{\sigma_{\mathcal{C}}+\delta},
    \label{eq:calibration_advantage}
\end{equation}
where $\mu_{\mathcal{C}}=M^{-1}\sum_{j=1}^{M}s_j$, $\sigma_{\mathcal{C}}=[M^{-1}\sum_{j=1}^{M}(s_j-\mu_{\mathcal{C}})^2]^{1/2}$, and $\delta>0$ ensures numerical stability.
The real-only calibration protocol uses global statistics over $\mathcal{C}$. 
The outcome weight $A_j$ is treated as constant during differentiation through the stop-gradient operator $\operatorname{sg}(\cdot)$.

The standardized outcomes define the policy-aware calibration surrogate
\begin{equation}
    \widetilde{\mathcal{J}}_{\mathrm{PA}}(\beta)
    :=-\frac{1}{M}\sum_{j=1}^{M}
    \operatorname{sg}(A_j)f_j(\beta).
    \label{eq:planning_aligned_surrogate}
\end{equation}
Combining Eqs.~\ref{eq:pairwise_influence} and~\ref{eq:planning_aligned_surrogate} gives
\begin{equation}
\begin{aligned}
    \left.\frac{d\widetilde{\mathcal{J}}_{\mathrm{PA}}
    (\hat{\beta}_{\eta,i})}{d\eta}\right|_{\eta=0}
    &=-\frac{1}{M}\sum_{j=1}^{M}A_j I_{ij} 
    &=\frac{1}{M}\sum_{j=1}^{M}A_j g_{ij}.
\end{aligned}
    \label{eq:surrogate_influence}
\end{equation}
Defining
\begin{equation}
    \psi_i:=\sum_{j=1}^{M}A_jg_{ij},
    \qquad
    \psi_i
    =M\left.\frac{d\widetilde{\mathcal{J}}_{\mathrm{PA}}
    (\hat{\beta}_{\eta,i})}{d\eta}\right|_{\eta=0},
    \label{eq:planning_aligned_influence}
\end{equation}
establishes $\psi_i$ as the candidate's local influence on the policy-aware calibration surrogate. The factor $M$ is common to all candidates and does not affect their ranking. 
Positive and negative $A_j$ distinguish above- and below-average policy outcomes. Consequently, $\psi_i$ favors candidates whose local training effects follow gradient structure associated with stronger policy behavior and oppose structure associated with weaker behavior. 

\subsection{Scalable Valuation and Selection}
\label{sec:influence_selection}

Computing $H^{-1}$ and all pairwise influences in Eq.~\ref{eq:pairwise_influence} is infeasible for a high-dimensional policy. PAD draws a fixed Gaussian projection $R\in\mathbb{R}^{P\times p}$ with $R_{ab}\sim\mathcal{N}(0,1/p)$ and represents candidate and calibration gradients as
\begin{equation}
    \phi_i:=R^{\top}q_i,
    \qquad
    \phi_j^{\mathrm{cal}}:=R^{\top}q_j^{\mathrm{cal}}
    \in\mathbb{R}^{p}.
    \label{eq:projected_gradients}
\end{equation}
For DiffusionDrive, $\xi=(t,\epsilon)$ indexes the timestep and noise used to perturb its trajectory anchors, and $\ell_{\mathrm{pl}}$ is the native multimodal trajectory loss. With $K$ Monte Carlo draws, Eq.~\ref{eq:projected_gradients} is estimated by
\begin{equation}
    \phi_i\approx\frac{1}{K}\sum_{k=1}^{K}
    R^{\top}\nabla_{\beta}
    \mathcal{L}_{\mathrm{traj}}\!\left(
    F_{\hat{\theta}}(o_i;t_k,\epsilon_k),\tau_i^{\star}
    \right).
    \label{eq:diffusiondrive_projected_gradient}
\end{equation}
Other policies use the same interface with their native supervised trajectory objectives. Policy-specific implementations are provided in the supplementary material.

Let $\mathcal{D}_{h}=\{z_{\ell}^{h}\}_{\ell=1}^{N_h}\subseteq\mathcal{D}_{\mathrm{real}}$ and $\Phi_h=[\phi_1^{h},\ldots,\phi_{N_h}^{h}]^{\top}$. The implementation forms the damped projected gradient-covariance preconditioner
\begin{equation}
    \widehat{G}
    :=\frac{1}{N_h}\Phi_h^{\top}\Phi_h+\lambda I_p,
    \qquad \lambda>0.
    \label{eq:projected_preconditioner}
\end{equation}
$\widehat{G}$ is an empirical projected gradient second moment with damping, not the exact Hessian $H$. 
PAD uses it as a TRAK-style approximation~\cite{park2023trak} to the local inverse training geometry:
\begin{equation}
    g_{ij}\approx
    (\phi_j^{\mathrm{cal}})^{\top}\widehat{G}^{-1}\phi_i.
    \label{eq:projected_helpfulness}
\end{equation}

Let $\Phi_{\mathrm{cal}}=[\phi_1^{\mathrm{cal}},\ldots,\phi_M^{\mathrm{cal}}]^{\top}$ and $\boldsymbol{A}=[A_1,\ldots,A_M]^{\top}$. The policy-aware calibration direction is
\begin{equation}
    c_{\mathrm{PA}}
    :=\Phi_{\mathrm{cal}}^{\top}\boldsymbol{A}
    =\sum_{j=1}^{M}A_j\phi_j^{\mathrm{cal}}
    =-M R^{\top}\nabla_{\beta}
    \widetilde{\mathcal{J}}_{\mathrm{PA}}(\hat{\beta}).
    \label{eq:planning_aligned_direction}
\end{equation}
Consequently, the projected estimate of Eq.~\ref{eq:planning_aligned_influence} is
\begin{equation}
    \psi_i\approx
    \phi_i^{\top}\widehat{G}^{-1}c_{\mathrm{PA}}.
    \label{eq:pair_influence_score}
\end{equation}
The subscript $\mathrm{PA}$ denotes policy awareness. 
Rather than materializing the $N\times M$ pairwise influence matrix, PAD solves $\widehat{G}u_{\mathrm{PA}}=c_{\mathrm{PA}}$ once and computes $\psi_i\approx\phi_i^{\top}u_{\mathrm{PA}}$ for every candidate.

Finally, for a subset $\mathcal{S}\subseteq\mathcal{D}_{\mathrm{pool}}$, define the joint perturbation
\begin{equation}
    \hat{\beta}_{\kappa,\mathcal{S}}
    :=\underset{\beta}{\arg\min}\;
    \left[L_{\mathrm{real}}(\beta)
    +\kappa\sum_{z_i\in\mathcal{S}}\bar{\ell}(z_i;\beta)\right],
    \quad \kappa>0.
    \label{eq:set_perturbation}
\end{equation}
For small $\kappa$, first-order superposition gives
\begin{equation}
    \Delta\widetilde{\mathcal{J}}_{\mathrm{PA}}(\mathcal{S})
    :=\widetilde{\mathcal{J}}_{\mathrm{PA}}
    (\hat{\beta}_{\kappa,\mathcal{S}})
    -\widetilde{\mathcal{J}}_{\mathrm{PA}}(\hat{\beta})
    \approx\frac{\kappa}{M}
    \sum_{z_i\in\mathcal{S}}\psi_i,
    \label{eq:set_influence_approximation}
\end{equation}
where higher-order interactions among candidates are omitted. Since $\kappa/M$ is positive and common to all feasible subsets, maximizing this approximation under $|\mathcal{S}|=B$ yields
\begin{equation}
    \mathcal{S}(B)
    :=\underset{\mathcal{S}\in\Omega_B}{\arg\max}\;
    \sum_{z_i\in\mathcal{S}}\psi_i
    =\underset{z_i\in\mathcal{D}_{\mathrm{pool}}}{\operatorname{TopB}}\;\psi_i.
    \label{eq:budgeted_selection}
\end{equation}
The selected demonstrations are combined with $\mathcal{D}_{\mathrm{real}}$, and the complete policy is retrained using its native architecture, objective, and training recipe.

\begin{figure}
    \centering
    \includegraphics[width=\linewidth]{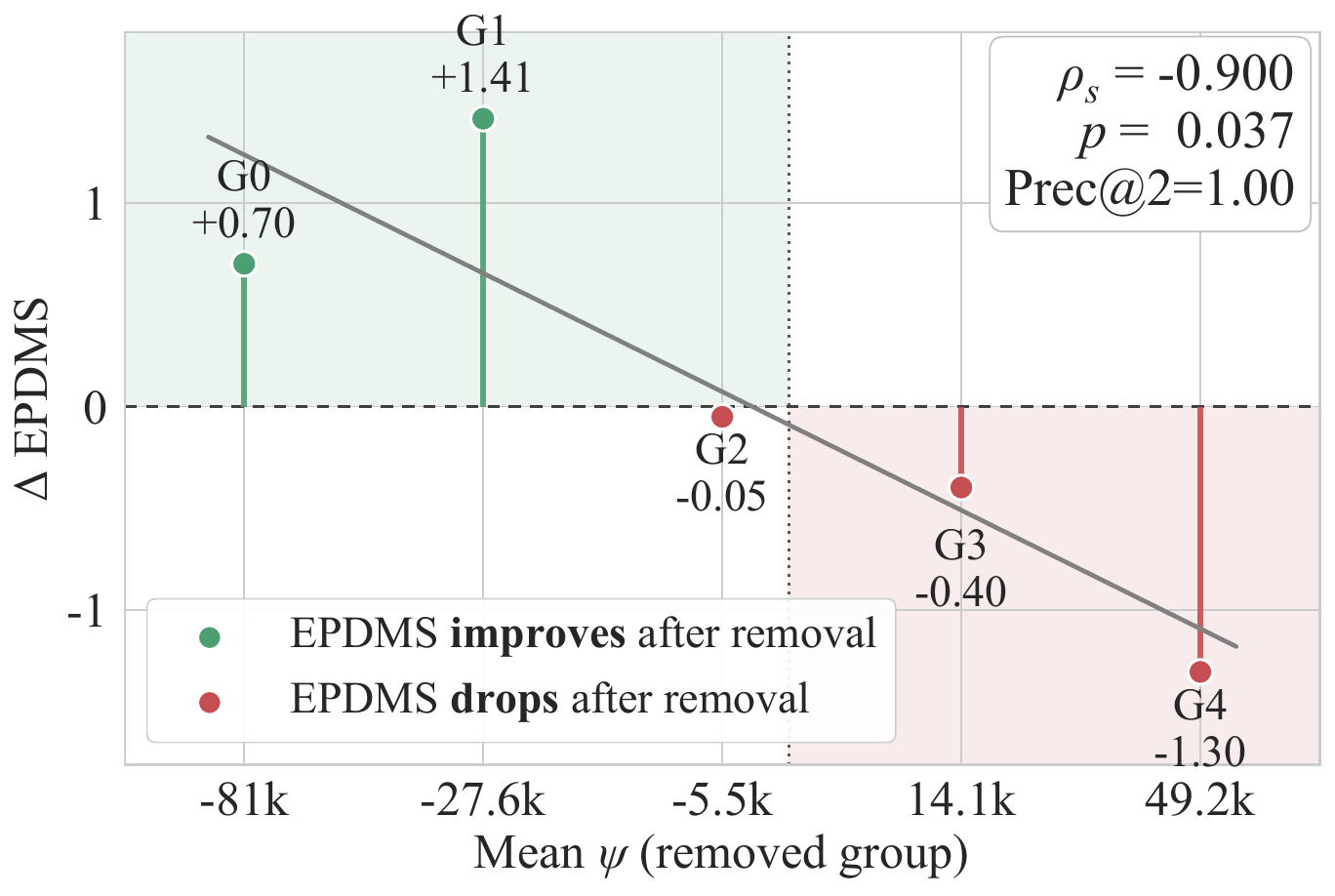}
    \caption{Removal-based validation of $\psi$. Each point measures the EPDMS change after removing one score-ranked group and retraining the planner.}
    \label{fig:influence_fidelity}
\end{figure}

\begin{table*}[t]
    \centering
    \renewcommand{\arraystretch}{1.08}
        \begin{tabular}{
            l!{\thinvrule}
            l!{\thinvrule}
            cccc!{\thinvrule}
            ccccc!{\thinvrule}
            c
        }
            \toprule

            \textbf{Method}
            & \textbf{Data Source}
            & \textbf{NC}$\uparrow$
            & \textbf{DAC}$\uparrow$
            & \textbf{DDC}$\uparrow$
            & \textbf{TLC}$\uparrow$
            & \textbf{EP}$\uparrow$
            & \textbf{TTC}$\uparrow$
            & \textbf{LK}$\uparrow$
            & \textbf{HC}$\uparrow$
            & \textbf{EC}$\uparrow$
            & \textbf{EPDMS}$\uparrow$ \\
            \midrule

            Human
            & -
            & 100
            & 100
            & 99.8
            & 100
            & 87.4
            & 100
            & 100
            & 98.1
            & 90.1
            & 90.3 \\
            
            Vanilla
            & Real
            & 98.2
            & 96.3
            & 99.5
            & 99.8
            & 87.6
            & 97.3
            & 97.2
            & 98.3
            & \textbf{88.0}
            & 84.8 \\


            SimScale
            & Real+Syn@PDM
            & 98.5
            & \textbf{97.1}
            & 99.6
            & 99.8
            & 87.5
            & 97.7
            & \textbf{97.9}
            & 98.3
            & 87.4
            & 85.9 \\


            SimScale
            & Real+Syn@Full
            & 98.1
            & 95.3
            & 99.5
            & 99.8
            & 96.5
            & 97.5
            & 97.4
            & 98.3
            & 76.5
            & 85.5 \\


            AutoScale
            & Real+Syn@100k
            & 98.5
            & 97.0
            & 99.6
            & 99.8
            & 87.5
            & 97.7
            & 97.4
            & 98.3
            & 87.9
            & 85.9 \\


            Random
            & Real+Syn@0.25
            & 98.1
            & 95.4
            & 99.5
            & 99.8
            & \textbf{96.7}
            & 97.3
            & 97.0
            & 98.3
            & 79.0
            & 85.7 \\


            \rowcolor{oursblue}
            PAD
            & Real+Syn@0.25
            & 98.3
            & 96.7
            & \textbf{99.6}
            & \textbf{99.8}
            & 96.6
            & 97.5
            & 97.4
            & \textbf{98.3}
            & 87.9
            & \textbf{88.3}\improvement{+3.5} \\

            \bottomrule
        \end{tabular}%
    \caption{Performance on NavTest Leaderboard. PAD retains top-ranked $25\%$ of the candidate pool.}
    \label{tab:main_results}
\end{table*}

\begin{table*}[t]
    \centering
    \renewcommand{\arraystretch}{1.2}
    \small
        \begin{tabular}{
            l!{\thinvrule}
            l!{\thinvrule}
            c!{\thinvrule}
            cccc!{\thinvrule}
            ccccc!{\thinvrule}
            c
        }
            \toprule

            \textbf{Method}
            & \textbf{Data Source}
            & \textbf{Stage}
            & \textbf{NC}$\uparrow$
            & \textbf{DAC}$\uparrow$
            & \textbf{DDC}$\uparrow$
            & \textbf{TLC}$\uparrow$
            & \textbf{EP}$\uparrow$
            & \textbf{TTC}$\uparrow$
            & \textbf{LK}$\uparrow$
            & \textbf{HC}$\uparrow$
            & \textbf{EC}$\uparrow$
            & \textbf{EPDMS}$\uparrow$ \\
            \midrule

            \multirow{2}{*}{Vanilla}
            & \multirow{2}{*}{Real}
            & S 1
            & 96.6
            & 85.6
            & 99.0
            & 99.3
            & 84.4
            & 95.3
            & 95.1
            & 97.6
            & 81.8
            & \multirow{2}{*}{27.9} \\

            &
            & S 2
            & 80.5
            & 72.2
            & 84.8
            & 98.7
            & 86.7
            & 76.7
            & 45.9
            & 95.3
            & 71.8
            & \\

            \methodrule

            \multirow{2}{*}{SimScale}
            & \multirow{2}{*}{Real+Syn@PDM}
            & S 1
            & 97.9
            & 89.6
            & 99.6
            & 99.3
            & 83.8
            & 96.9
            & 98.2
            & 97.6
            & 76.4
            & \multirow{2}{*}{31.7} \\

            &
            & S 2
            & 84.0
            & 68.7
            & 92.7
            & 98.0
            & 93.3
            & 77.3
            & 63.5
            & 95.2
            & 40.1
            & \\

            \methodrule

            \multirow{2}{*}{SimScale}
            & \multirow{2}{*}{Real+Syn@Full}
            & S 1
            & 96.8
            & 87.1
            & 99.3
            & 99.8
            & 84.4
            & 96.7
            & 98.4
            & 97.6
            & 78.2
            & \multirow{2}{*}{31.3} \\

            &
            & S 2
            & 84.2
            & 70.8
            & 90.8
            & 98.6
            & 91.2
            & 77.3
            & 61.4
            & 96.0
            & 39.2
            & \\

            \methodrule

            \multirow{2}{*}{AutoScale$^{\dagger}$}
            & \multirow{2}{*}{Real+Syn@100k}
            & S 1
            & 97.9
            & 89.1
            & 99.2
            & 99.3
            & 84.3
            & 97.1
            & 98.0
            & 97.8
            & 77.8
            & \multirow{2}{*}{33.5} \\

            &
            & S 2
            & 83.4
            & 72.7
            & 91.6
            & 98.8
            & 90.0
            & 77.4
            & 61.2
            & 94.6
            & 46.1
            & \\

            \methodrule

            \multirow{2}{*}{Random}
            & \multirow{2}{*}{Real+Syn@0.25}
            & S 1
            & 96.7
            & 84.4
            & 98.8
            & 99.3
            & 84.0
            & 95.1
            & 96.7
            & 97.6
            & 67.6
            & \multirow{2}{*}{29.2} \\

            &
            & S 2
            & 83.8
            & 72.8
            & 92.9
            & 98.9
            & 88.3
            & 78.8
            & 57.7
            & 95.2
            & 40.0
            & \\

            \methodrule

            \rowcolor{oursblue}
            &
            & S 1
            & 97.8
            & 89.8
            & 99.2
            & 99.3
            & 84.5
            & 96.4
            & 96.9
            & 97.8
            & 75.1
            & \\
            
            \rowcolor{oursblue}
            \multirow{-2}{*}{\textbf{PAD}}
            & \multirow{-2}{*}{Real+Syn@0.25}
            & S 2
            & 83.3
            & 69.7
            & 91.4
            & 98.4
            & 91.0
            & 77.1
            & 59.4
            & 95.1
            & 46.7
            & \multirow{-2}{*}{\textbf{32.9}\gain{+5.0}} \\

            \bottomrule
        \end{tabular}%
    \caption{Performance on NavHard Leaderboard. Per-stage metrics are reported for both stages, and EPDMS denotes the combined score. $^{\dagger}$AutoScale results are taken from the original paper, as its code and model weights are not publicly available.}
    \label{tab:stagewise_results} 
\end{table*}

\section{Experiments}
\label{Experiments}

\subsection{Experimental Setup}
\label{sec:experimental_setup}

\paragraph{Benchmarks and Metrics.}
We evaluate on NAVSIM v2~\cite{Cao2025CORL} using real \texttt{navtrain} demonstrations for its established benchmark protocol and challenging pseudo-closed-loop evaluation, and the SimScale synthetic pool~\cite{tian2026simscale} generated by PDM and recovery pseudo-experts over five rounds. 
A held-out real split forms the calibration set $\mathcal{C}$, while NavTest and NavHard are reserved exclusively for final evaluation. 
Planning performance is quantified by the Extended Predictive Driver Model Score (EPDMS), which combines multiplicative safety and compliance penalties with weighted measures of progress, time to collision, etc.

\paragraph{Policies and Implementation.}
We consider three end-to-end policies, spanning the diffusion-based DiffusionDrive~\cite{liao2025diffusiondrive}, the regression-based LTF~\cite{chitta2022transfuser,dauner2024navsim}, and the flow-matching-based MeanFuser~\cite{wang2026meanfuser}. 
All policies receive the same camera-only input, combining a stitched current front view with the ego command, velocity, and acceleration, and predict a four-second trajectory at $0.5$-second intervals.
Each configuration is trained with its native objective for 100 epochs on four NVIDIA A800 GPUs and evaluated on four NVIDIA L40 GPUs.
Architectural and optimization details are provided in the supplementary material.

\subsection{Valuation Fidelity}
\label{sec:influence_fidelity}

\paragraph{Removal-Based Verification.}

\begin{figure*}[t]
    \centering
    \includegraphics[width=\linewidth]{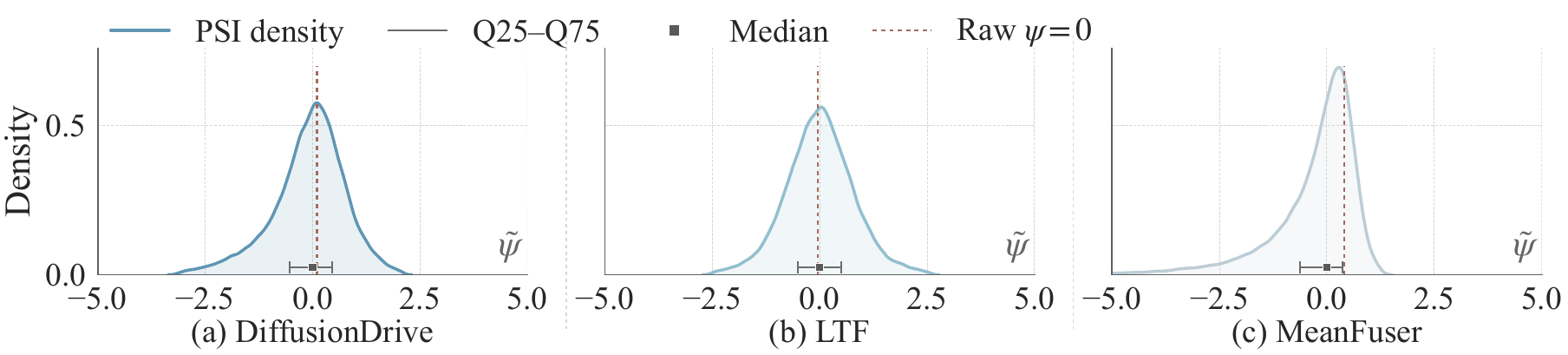}
    \caption{Policy-specific PAD values for the same synthetic candidate pool across DiffusionDrive, LTF, and MeanFuser. Values are standardized within each policy.}
    \label{fig:psi_distribution}
\end{figure*}

To determine whether $\psi$ predicts realized training utility, a controlled group-removal experiment is conducted on DiffusionDrive.
The synthetic demonstrations are sorted by $\psi$ and partitioned into five equal-sized groups $G_0,\ldots,G_4$. 
Each group is removed independently, and the policy is retrained from the same base checkpoint for 20 epochs.
A matched run retaining the full synthetic pool serves as the reference. For group $G_k$, the removal effect is
\begin{equation}
    \Delta_k
    = \operatorname{EPDMS}(\pi_{-G_k})
    - \operatorname{EPDMS}(\pi_{\mathrm{full}}).
    \label{eq:removal_effect}
\end{equation}
If $\psi$ faithfully ranks training utility, $\Delta_k$ should decrease with the mean value of the removed group.

\paragraph{Removal Effects Prediction.}
Figure~\ref{fig:influence_fidelity} shows a clear directional correspondence between predicted influence and measured removal effects. 
The full-pool reference achieves an EPDMS of $84.31$. 
Removing the two lowest-valued groups increases EPDMS by $0.70$ and $1.41$, while removing each of the three higher-valued groups reduces performance. 
Removing the highest-valued group causes the largest decrease of $1.30$ points. 
Across all groups, mean $\psi$ and the removal effect exhibit a strong negative rank correlation of $\rho_s=-0.900$ with $p=0.0374$. 
Both lowest-valued groups are correctly identified as harmful under the removal criterion, yielding $\mathrm{Prec@2}=1.00$. 
These results establish the directional fidelity of PAD valuation at the group level.

\subsection{NavTest and NavHard Benchmarks}
\label{sec:data_efficient_curation}

\paragraph{NavTest Leaderboard.}
Table~\ref{tab:main_results} compares DiffusionDrive under different synthetic-data configurations.
Retaining the top-ranked $25\%$ of the candidate pool, PAD achieves an EPDMS of $88.3$.
This exceeds the real-only model by $3.5$ points and the budget-matched random baseline by $2.6$ points.
PAD also outperforms SimScale trained with PDM synthetic data and the complete synthetic pool by $2.4$ and $2.8$ points, respectively. 
The gains therefore arise from policy-aware valuation rather than synthetic-data volume alone.

\paragraph{NavHard Leaderboard.}
Table~\ref{tab:stagewise_results} evaluates the same configurations under the two-stage NavHard protocol. 
PAD reaches a combined EPDMS of $32.9$, exceeding real-only training by $5.0$ points and budget-matched random selection by $3.7$ points. 
It also surpasses SimScale trained with PDM data and the complete synthetic pool by $1.2$ and $1.6$ points, respectively. PAD thus achieves the strongest aggregate performance among the controlled baselines.

\subsection{Policy Specificity and Generalization}
\label{sec:cross_planner_generalization}

\paragraph{Policy-Specific Valuation.}
We first examine whether the data value transfers across policies. 
For the same 208,413 synthetic demonstrations, $\psi$ is recomputed independently with each planner's parameters and planning objective. 
Figure~\ref{fig:psi_distribution} reports the standardized score $\widetilde{\psi}=(\psi-\operatorname{median}(\psi))/\operatorname{IQR}(\psi)$, which differ substantially in shape and concentration. 
Pairwise Spearman correlations range from $0.026$ to $0.107$, while top-$25\%$ Jaccard overlap ranges from $0.146$ to $0.155$. 
Data value is therefore not an intrinsic property of a demonstration, but is conditioned on the policy that learns from it.

\begin{figure}[t]
    \centering
    \subfloat[Regression severity under group removal.]{
        \includegraphics[width=0.45\columnwidth]
        {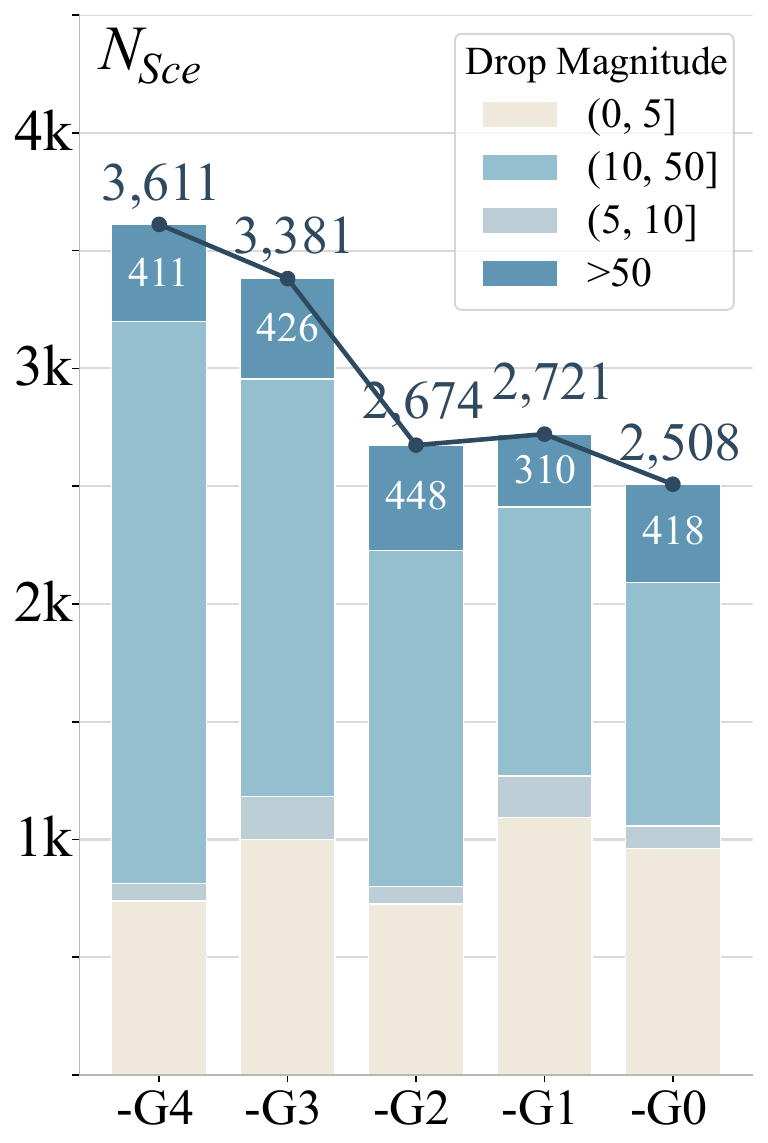}
        \label{fig:regression_severity}
    }
    \hfill
    \subfloat[Regression breadth across policies.]{
        \includegraphics[width=0.45\columnwidth]
        {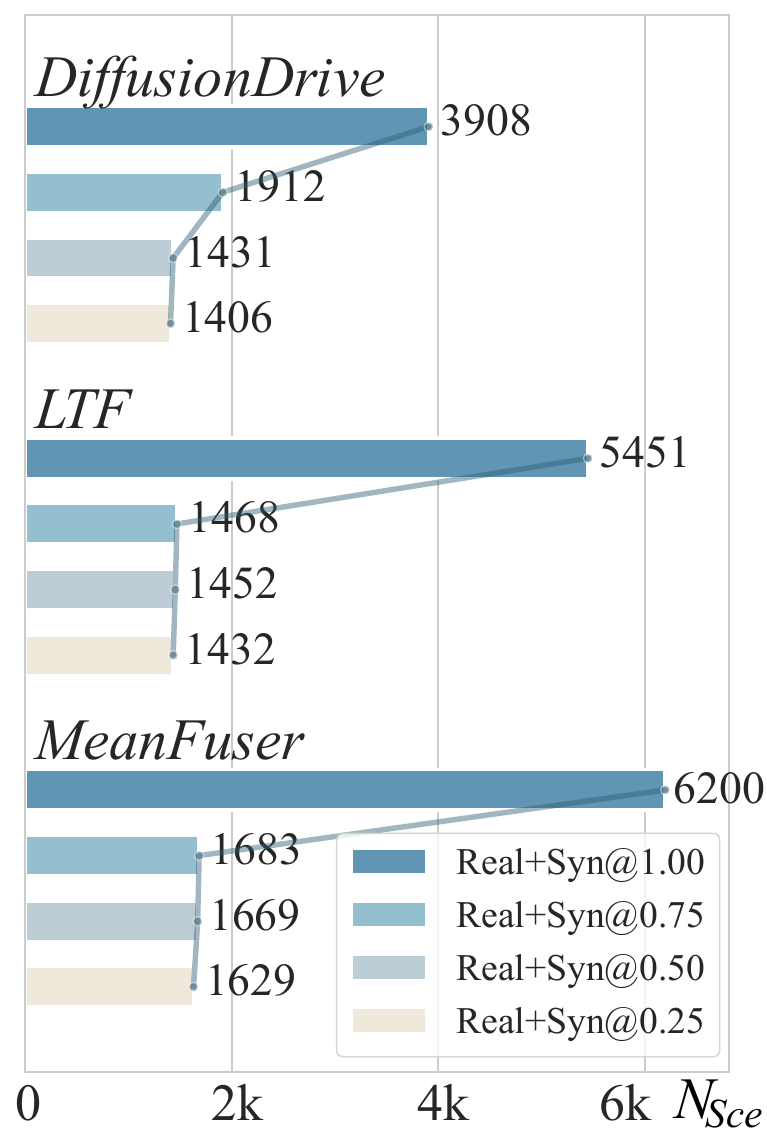}
        \label{fig:regression_breadth}
    }

    \caption{Scene-level regression exposure on NavTest.}
    \label{fig:scenario_reliability}
\end{figure}

\paragraph{Cross-Planner Benchmark.}
Table~\ref{tab:cross_planner_results} evaluates PAD on diffusion-based, regression-based, and flow-matching-based policies. 
At a $25\%$ synthetic-data budget, PAD reaches EPDMS scores of $88.3$, $85.6$, and $86.7$ for DiffusionDrive, LTF, and MeanFuser, respectively. 
Each result exceeds both the corresponding real-only planner and its full-data SimScale counterpart. 
Across the evaluated budgets, the best PAD configuration improves over full-pool SimScale by $2.4$ points on DiffusionDrive, $1.5$ points on LTF, and $2.3$ points on MeanFuser. 
The consistent improvement across distinct training objectives supports the general applicability of the proposed valuation framework.

\begin{table*}[t]
    \centering
    \renewcommand{\arraystretch}{1.08}
    \small
        \begin{tabular*}{\textwidth}{@{\extracolsep{\fill}}
            l!{\thinvrule}
            l!{\thinvrule}
            cccc!{\thinvrule}
            ccccc!{\thinvrule}
            c@{}
        }
            \toprule

            \textbf{Method}
            & \textbf{Data Source}
            & \textbf{NC}$\uparrow$
            & \textbf{DAC}$\uparrow$
            & \textbf{DDC}$\uparrow$
            & \textbf{TLC}$\uparrow$
            & \textbf{EP}$\uparrow$
            & \textbf{TTC}$\uparrow$
            & \textbf{LK}$\uparrow$
            & \textbf{HC}$\uparrow$
            & \textbf{EC}$\uparrow$
            & \textbf{EPDMS}$\uparrow$ \\
            \midrule

            \rowcolor{groupgray}
            \multicolumn{12}{c}{
                \textit{Diffusion-Based Planner (CVPR 2025)}
            } \\
            \rowrule

            DiffusionDrive
            & Real
            & 98.2
            & 96.3
            & 99.5
            & 99.8
            & 87.6
            & 97.3
            & 97.2
            & 98.3
            & \textbf{88.0}
            & 84.8 \\

            SimScale
            & Real+Syn
            & \textbf{98.5}
            & \textbf{97.1}
            & 99.6
            & 99.8
            & 87.5
            & \textbf{97.7}
            & \textbf{97.9}
            & 98.3
            & 87.4
            & 85.9\improvement{+1.1} \\

            PAD
            & Real+Syn@0.25
            & 98.3
            & 96.7
            & \textbf{99.6}
            & \textbf{99.8}
            & 96.6
            & 97.5
            & 97.4
            & \textbf{98.3}
            & 87.9
            & \textbf{88.3}\improvement{+3.5} \\

            PAD
            & Real+Syn@0.50
            & 98.2
            & 96.5
            & 99.6
            & 99.8
            & 96.8
            & 97.5
            & 97.4
            & 98.3
            & 87.5
            & 88.0\improvement{+3.1} \\

            PAD
            & Real+Syn@0.75
            & 98.0
            & 96.8
            & 99.6
            & 99.8
            & \textbf{96.9}
            & 97.4
            & 97.5
            & 98.3
            & 80.1
            & 87.3\improvement{+2.4} \\

            \midrule

            \rowcolor{groupgray}
            \multicolumn{12}{c}{
                \textit{Regression-Based Planner (NeurIPS 2024)}
            } \\
            \rowrule

            LTF
            & Real
            & 96.5
            & 90.2
            & 98.2
            & 99.7
            & 95.1
            & 96.0
            & 93.7
            & 98.3
            & 87.1
            & 79.1 \\

            SimScale
            & Real+Syn
            & \textbf{98.3}
            & \textbf{95.6}
            & \textbf{99.6}
            & 99.8
            & 87.1
            & \textbf{97.5}
            & \textbf{97.2}
            & 98.3
            & \textbf{88.2}
            & 84.4\improvement{+5.3} \\

            PAD
            & Real+Syn@0.25
            & 98.0
            & 94.5
            & 99.4
            & \textbf{99.9}
            & 96.5
            & 97.2
            & 96.3
            & 98.3
            & 87.5
            & 85.6\improvement{+6.5} \\
            
            PAD
            & Real+Syn@0.50
            & 98.0
            & 94.0
            & 99.5
            & 99.8
            & \textbf{96.6}
            & 97.0
            & 96.7
            & 98.3
            & 87.9
            & 85.1\improvement{+6.0} \\

            PAD
            & Real+Syn@0.75
            & 98.0
            & 94.7
            & 99.4
            & 99.8
            & 96.5
            & 97.3
            & 96.9
            & \textbf{98.4}
            & 87.1
            & \textbf{85.9}\improvement{+6.8} \\

            \midrule

            \rowcolor{groupgray}
            \multicolumn{12}{c}{
                \textit{Flow-Matching-Based Planner (CVPR 2026)}
            } \\
            \rowrule

            MeanFuser
            & Real
            & 97.6
            & 93.6
            & 99.2
            & 99.8
            & 96.4
            & 96.7
            & 96.2
            & 98.3
            & 87.7
            & 84.2 \\

            SimScale
            & Real+Syn
            & 97.9
            & \textbf{96.5}
            & 99.5
            & 99.8
            & 87.4
            & 97.2
            & \textbf{97.4}
            & 98.3
            & \textbf{87.9}
            & 84.8\improvement{+0.5} \\

            PAD
            & Real+Syn@0.25
            & 97.8
            & 95.6
            & 99.5
            & 99.8
            & 96.5
            & 97.0
            & 97.0
            & 98.3
            & 87.8
            & 86.7\improvement{+2.5} \\

            PAD
            & Real+Syn@0.50
            & 97.9
            & 95.6
            & 99.5
            & 99.8
            & 96.5
            & \textbf{97.2}
            & 97.0
            & 98.3
            & 87.6
            & 86.7\improvement{+2.5} \\

            PAD
            & Real+Syn@0.75
            & \textbf{97.9}
            & 95.9
            & \textbf{99.6}
            & \textbf{99.8}
            & \textbf{96.5}
            & 97.1
            & 97.2
            & \textbf{98.3}
            & 87.5
            & \textbf{87.1}\improvement{+2.9} \\

            \bottomrule
        \end{tabular*}%
    \caption{NavTest performance across policy families and selection budgets. PAD independently estimates $\psi$ for each policy and retains the top-ranked $25\%$, $50\%$, or $75\%$ of the common candidate pool.}
    \label{tab:cross_planner_results}
\end{table*}

\begin{figure*}
    \centering
    \includegraphics[width=\linewidth]{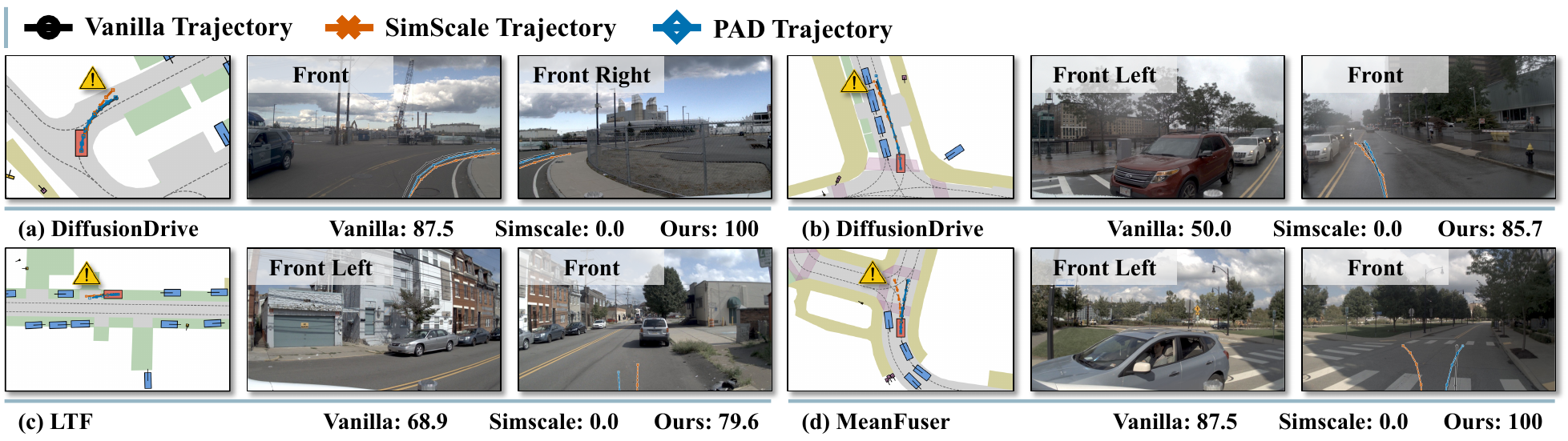}
    \caption{Representative NavTest trajectories across policy families. Full-pool synthetic training produces zero-score failures, whereas PAD retains feasible, higher-scoring predictions.}
    \label{fig:qualitative_analysis}
\end{figure*}

\paragraph{Budget Sensitivity.}
The optimal selection budget varies across policies. DiffusionDrive performs best at $25\%$, after which EPDMS decreases from $88.3$ to $88.0$ and $87.3$. LTF varies non-monotonically and peaks at $85.9$ with $75\%$, while MeanFuser improves from $86.7$ at $25\%$ to $87.1$ at $75\%$. Nevertheless, every PAD budget outperforms full-pool SimScale for all three policies. These results show that additional data are not uniformly beneficial and that the optimal retention ratio depends on the target policy. Complete NavHard results are provided in the supplementary material.

\subsection{Scenario-Level Reliability}
\label{sec:scenario_reliability}

\paragraph{Regression Metrics.}
For a policy configuration $b$ and reference $r$, let $q_n^{(b)}$ and $q_n^{(r)}$ denote their EPDMS values on scene $n$, and define $\delta_n^{(b\mid r)}=q_n^{(b)}-q_n^{(r)}$. 
A scene is degraded when $\delta_n^{(b\mid r)}<0$. Its severity is $d_n^{(b\mid r)}=\max(0,-\delta_n^{(b\mid r)})$, and the regression footprint is $R^{(b\mid r)}=N^{-1}\sum_{n=1}^{N}\mathbf{1}[\delta_n^{(b\mid r)}<0]$. 
We define a catastrophic regression by $\delta_n^{(b\mid r)}<-50$. The removal analysis uses the full-pool policy as $r$, while the budget analysis uses the corresponding real-only policy.

\paragraph{PAD Reduces Regression Exposure.}
All augmented configurations improve mean EPDMS, yet scene-level regressions remain. 
In Fig.~\ref{fig:scenario_reliability}(a), removing the highest-valued group $G_4$ produces $3{,}611$ degraded scenes and $411$ catastrophic regressions. Removing either of the two lowest-valued groups produces $2{,}508$ to $2{,}721$ degraded scenes, with catastrophic-regression counts ranging from $310$ to $418$. 
High-$\psi$ demonstrations therefore support a narrower overall regression footprint. 
In Fig.~\ref{fig:scenario_reliability}(b), PAD at $25\%$ limits degraded-scene shares to $11.6\%$, $11.8\%$, and $13.4\%$ for DiffusionDrive, LTF, and MeanFuser, compared with $32.2\%$, $44.9\%$, and $51.0\%$ under full-pool training. 
These shares remain below $15.8\%$ at every PAD budget. 

\paragraph{Qualitative Analysis.}
Figure~\ref{fig:qualitative_analysis} examines representative failures across three policy families. 
In the two DiffusionDrive scenes, PAD improves the real-only EPDMS from $87.5$ to $100$ on a curved roadway and from $50.0$ to $85.7$ at an intersection with cross traffic, while full-pool training yields zero scores in both cases. 
The same pattern extends to LTF and MeanFuser. PAD raises EPDMS from $68.9$ to $79.6$ in a dense urban scene and from $87.5$ to $100$ at an intersection, whereas both full-pool policies again score zero. 
Across these settings, PAD preserves route-consistent trajectories while avoiding the severe local failures introduced by indiscriminate augmentation. 
Further qualitative results are provided in the supplementary material.

\section{Conclusion}

This paper introduced PAD, a policy-aware framework that defines the value of a driving demonstration by its downstream utility to the policy that learns from it. 
By combining independent scene-level policy outcomes with projected training gradients and regularized local geometry, PAD converts otherwise intractable retraining effects into scalable per-demonstration valuations. 
The results show that these values track realized training effects and enable policy-specific curation that consistently improves driving performance across distinct policy families while substantially reducing scene-level regression exposure. 
These findings establish that data utility is inherently policy dependent and that effective scaling requires optimizing data composition rather than volume alone. 
The current study is limited by its local first-order approximation and its evaluation on synthetic candidates within NAVSIM. Future work will model training-stage dependence and higher-order interactions, and extend validation to heterogeneous real-world data and broader closed-loop conditions. 

\bibliography{aaai2027}


\clearpage
\section*{Supplementary Material}

\begingroup
\renewcommand{\arraystretch}{1.28}
\noindent
\begin{tabularx}{\columnwidth}{@{}lXr@{}}
\textbf{A.} & \textbf{Method Derivations}\dotfill
& \textbf{\pageref{app:method}} \\
& A.1. Local Influence\dotfill
& \pageref{app:local_influence} \\
& A.2. Policy-Aware Estimation\dotfill
& \pageref{app:projected_estimation} \\
& A.3. Policy Instantiations\dotfill
& \pageref{app:policy_instantiations} \\

\textbf{B.} & \textbf{Extended Results}\dotfill
& \textbf{\pageref{app:extended_results}} \\
& B.1. Valuation Fidelity\dotfill
& \pageref{app:valuation_fidelity} \\
& B.2. Performance Across Policies\dotfill
& \pageref{app:cross_policy_results} \\

\textbf{C.} & \textbf{Experimental Details}\dotfill
& \textbf{\pageref{app:experimental_details}} \\
& C.1. Policy Configurations\dotfill
& \pageref{app:policy_configurations} \\
& C.2. Calibration Protocol\dotfill
& \pageref{app:calibration_protocol} \\
& C.3. Training and Baselines\dotfill
& \pageref{app:training_baselines} \\
& C.4. Reproducibility\dotfill
& \pageref{app:reproducibility} \\

\textbf{D.} & \textbf{Qualitative Results}\dotfill
& \textbf{\pageref{app:qualitative}} \\
& D.1. Planning Comparisons\dotfill
& \pageref{app:planning_comparisons} \\
& D.2. Failure Cases\dotfill
& \pageref{app:failure_cases} \\

\end{tabularx}
\endgroup

\vspace{0.5em}
\noindent\rule{\columnwidth}{0.4pt}

\appendix
\numberwithin{equation}{section}

\section{Method Derivations}
\label{app:method}

PAD values a demonstration by its local effect on a policy-aware calibration surrogate. 
This section establishes that construction in three steps: the
exact response to an infinitesimal candidate perturbation, its scalable
projected approximation, and its policy-specific instantiation. We follow the notation and equations of the main paper.

The restricted parameter subspace is used only to estimate data value. Final curated retraining optimizes the complete policy with its native objective.

\subsection{Local Influence}
\label{app:local_influence}

Let $n=|\mathcal{D}_{\mathrm{real}}|$ and abbreviate
$\bar{\ell}_i(\beta)=\bar{\ell}(z_i;\beta)$. Adding candidate $z_i$ to the
normalized empirical objective gives
\begin{equation}
\begin{aligned}
L_{+i}(\beta)
&=\frac{1}{n+1}\left[
\sum_{z\in\mathcal{D}_{\mathrm{real}}}\bar{\ell}(z;\beta)
+\bar{\ell}_i(\beta)\right] \\
&=\frac{n}{n+1}\left[
L_{\mathrm{real}}(\beta)+\frac{1}{n}\bar{\ell}_i(\beta)\right].
\end{aligned}
\label{eq:app_add_one_equivalence}
\end{equation}
The leading factor does not change the minimizer. Hence, the perturbation in
Eq.~(4) of the main paper is a continuous relaxation of literal addition, with
$\eta=1/n$ recovering its relative weight. The infinitesimal regime isolates
the candidate's first-order effect around the real-only checkpoint.

To derive this effect, define the stationarity map
$F_i(\beta,\eta)=\nabla_{\beta}L_{\mathrm{real}}(\beta)
+\eta\nabla_{\beta}\bar{\ell}_i(\beta)$. Equation~(4) implies
$F_i(\hat{\beta}_{\eta,i},\eta)=0$. Assume that both losses are twice
continuously differentiable near $\hat{\beta}$, that
$\nabla_{\beta}L_{\mathrm{real}}(\hat{\beta})=0$, and that the restricted
Hessian $H$ defined in the main paper is nonsingular. The implicit function
theorem then yields a locally unique differentiable stationary branch
$\eta\mapsto\hat{\beta}_{\eta,i}$. Writing
$\dot{\beta}_{\eta,i}=d\hat{\beta}_{\eta,i}/d\eta$, differentiation before
setting $\eta=0$ gives
\begin{equation}
\left[
\nabla_{\beta}^{2}L_{\mathrm{real}}(\hat{\beta}_{\eta,i})
+\eta\nabla_{\beta}^{2}\bar{\ell}_i(\hat{\beta}_{\eta,i})
\right]\dot{\beta}_{\eta,i}
+\nabla_{\beta}\bar{\ell}_i(\hat{\beta}_{\eta,i})=0.
\label{eq:app_differentiated_stationarity}
\end{equation}
Evaluating Eq.~\eqref{eq:app_differentiated_stationarity} at zero produces the
parameter derivative in Eq.~(6) and the local expansion
\begin{equation}
\hat{\beta}_{\eta,i}
=\hat{\beta}-\eta H^{-1}q_i+o(\eta).
\label{eq:app_parameter_expansion}
\end{equation}

The nonsingularity assumption ensures that this response is unique; positive
definiteness supplies the local-minimum interpretation. Indeed, at a
twice-differentiable local minimum, $H\succeq0$. Write its eigendecomposition
as $H=U\operatorname{diag}(\lambda_1,\ldots,\lambda_P)U^{\top}$, where
$U=[u_1,\ldots,u_P]$. If $H$ is also nonsingular, then every $\lambda_r>0$ and
\begin{equation}
H^{-1}q_i
=\sum_{r=1}^{P}\frac{u_r^{\top}q_i}{\lambda_r}u_r.
\label{eq:app_spectral_response}
\end{equation}
This expression also exposes the numerical difficulty: directions with small
$\lambda_r$ amplify the corresponding gradient components. A nonsingular but
indefinite Hessian still defines the stationary derivative, but not a stable
minimum. Section~A.2 therefore replaces this exact local reference with damped
projected geometry.

Finally, apply a first-order expansion of the calibration loss in Eq.~(7):
\begin{equation}
\begin{aligned}
f_j(\hat{\beta}_{\eta,i})
&=f_j(\hat{\beta})
+(q_j^{\mathrm{cal}})^{\top}
(\hat{\beta}_{\eta,i}-\hat{\beta})
+o(\|\hat{\beta}_{\eta,i}-\hat{\beta}\|) \\
&=f_j(\hat{\beta})
-\eta(q_j^{\mathrm{cal}})^{\top}H^{-1}q_i
+o(\eta).
\end{aligned}
\label{eq:app_calibration_expansion}
\end{equation}
The coefficient of $\eta$ is the influence $I_{ij}$ in Eq.~(8); its negative
is the helpfulness $g_{ij}$. Hence, $g_{ij}>0$ predicts a local decrease in
calibration loss. The candidate is absent from $H$ because curvature is
evaluated at the unperturbed real-data objective. 
These identities characterize the first-order response of the restricted problem; they do not equate an infinitesimal effect with finite retraining. Upstream parameters remain trainable during final curated retraining. Section~A.2 converts the scene-wise responses into policy-aware data values.

\subsection{Policy-Aware Estimation}
\label{app:projected_estimation}

\paragraph{Policy-aware direction.}
Equations~(9)--(12) of the main paper condition the local calibration target
on the base policy's scene-level outcomes. The normalization in Eq.~(9) gives
\begin{equation}
\sum_{j=1}^{M}A_j=0,
\qquad
\sum_{j=1}^{M}A_j^2
=\frac{M\sigma_{\mathcal{C}}^2}
{(\sigma_{\mathcal{C}}+\delta)^2}.
\label{eq:app_advantage_properties}
\end{equation}
The stop-gradient operator then treats these coefficients as fixed during
differentiation.
The first identity makes the target contrastive: a gradient shared uniformly
across calibration scenes cancels, whereas gradient structure associated
differently with stronger and weaker policy outcomes remains. Define the
outcome-weighted aggregate as
\begin{equation}
q_{\mathrm{PA}}
:=\sum_{j=1}^{M}A_jq_j^{\mathrm{cal}}
=-M\nabla_{\beta}
\widetilde{\mathcal{J}}_{\mathrm{PA}}(\hat{\beta}).
\label{eq:app_unprojected_pa_direction}
\end{equation}
Combining Eq.~\eqref{eq:app_unprojected_pa_direction} with the parameter
response in Eq.~(6) of the main paper yields
\begin{equation}
M\left.
\frac{d\widetilde{\mathcal{J}}_{\mathrm{PA}}
(\hat{\beta}_{\eta,i})}{d\eta}
\right|_{\eta=0}
=q_{\mathrm{PA}}^{\top}H^{-1}q_i.
\label{eq:app_unprojected_pa_influence}
\end{equation}
The right-hand side is the value $\psi_i$ defined in Eq.~(12). Outcome
conditioning enters through $q_{\mathrm{PA}}$, whose weights $A_j$ depend on
the base policy's EPDMS. The full value is policy specific because
$q_j^{\mathrm{cal}}$, $H$, and $q_i$ are also computed from that policy and its
native trajectory objective. This construction is a supervised local surrogate
conditioned on policy outcomes, not a gradient of EPDMS itself.

\paragraph{Projected geometry.}
The exact bilinear form in Eq.~\eqref{eq:app_unprojected_pa_influence} is
intractable at the dimension of $\beta$. Equation~(13) therefore maps every
candidate, calibration, and geometry-set gradient through the same fixed random
projection $R$. For arbitrary fixed $a,b\in\mathbb{R}^{P}$, its Gaussian
normalization gives
\begin{equation}
\mathbb{E}_{R}[RR^{\top}]=I_P,
\qquad
\mathbb{E}_{R}
\left[(R^{\top}a)^{\top}(R^{\top}b)\right]=a^{\top}b.
\label{eq:app_projection_expectation}
\end{equation}
This identity motivates the shared projected representation, although it does
not make the final preconditioned score an unbiased estimator of exact
influence.

Following the main paper, let
$\mathcal{D}_{h}=\{z_{\ell}^{h}\}_{\ell=1}^{N_h}\subseteq\mathcal{D}_{\mathrm{real}}$
be the geometry set and let
$C_h=N_h^{-1}\sum_{\ell=1}^{N_h}q_\ell^h(q_\ell^h)^{\top}$ denote its
unprojected empirical gradient second moment. The
preconditioner in Eq.~(15) obeys
\begin{equation}
\widehat{G}
=R^{\top}C_hR+\lambda I_p.
\label{eq:app_projected_second_moment}
\end{equation}
It is therefore a damped projection of an empirical gradient second moment, not
an identity equating it with $H$. For any nonzero $v\in\mathbb{R}^{p}$,
\begin{equation}
v^{\top}\widehat{G}v
=\frac{1}{N_h}\lVert\Phi_hv\rVert_2^2
+\lambda\lVert v\rVert_2^2
\geq\lambda\lVert v\rVert_2^2>0.
\label{eq:app_preconditioner_pd}
\end{equation}
Hence, $\widehat{G}\succ0$ whenever $\lambda>0$, its inverse is unique, and
$\lVert\widehat{G}^{-1}\rVert_2\leq1/\lambda$. Damping thus bounds the
amplification caused by poorly supported projected directions. Relative to
exact influence, random compression and the substitution of
$\widehat{G}^{-1}$ for inverse Hessian geometry are approximation steps. Once
the projected quantities are fixed, the aggregation identities remain exact.

\paragraph{Compressed valuation.}
Let $\Phi_{\mathrm{pool}}=[\phi_1,\ldots,\phi_N]^{\top}$ and use
$\Phi_{\mathrm{cal}}$ and $\boldsymbol{A}$ from the main paper. Associativity
of the projected bilinear form gives the vector of all candidate scores as
\begin{equation}
\widehat{\boldsymbol{\psi}}
=\Phi_{\mathrm{pool}}\widehat{G}^{-1}
\Phi_{\mathrm{cal}}^{\top}\boldsymbol{A}
=\Phi_{\mathrm{pool}}u_{\mathrm{PA}},
\qquad
\widehat{G}u_{\mathrm{PA}}
=\Phi_{\mathrm{cal}}^{\top}\boldsymbol{A}.
\label{eq:app_vectorized_scores}
\end{equation}
The right-hand side of the linear system is precisely the compressed direction
$c_{\mathrm{PA}}$ in Eq.~(17). One solve therefore replaces all $NM$ pairwise
evaluations: calibration scenes are aggregated before preconditioning, and
each candidate requires only an inner product with $u_{\mathrm{PA}}$. This
reordering is exact for the projected estimator in Eq.~(18).

\paragraph{Budgeted selection.}
For the joint perturbation in Eq.~(19), differentiating its stationarity
condition as in Section~A.1 gives
\begin{equation}
\hat{\beta}_{\kappa,\mathcal{S}}
=\hat{\beta}
-\kappa H^{-1}\sum_{z_i\in\mathcal{S}}q_i
+o(\kappa).
\label{eq:app_set_parameter_expansion}
\end{equation}
Substitution into the policy-aware surrogate yields
\begin{equation}
M\left[
\widetilde{\mathcal{J}}_{\mathrm{PA}}
(\hat{\beta}_{\kappa,\mathcal{S}})
-\widetilde{\mathcal{J}}_{\mathrm{PA}}(\hat{\beta})
\right]
=\kappa\sum_{z_i\in\mathcal{S}}\psi_i+o(\kappa),
\label{eq:app_set_surrogate_expansion}
\end{equation}
which establishes the first-order additivity used in Eq.~(20). Under the fixed
cardinality constraint, this modular objective is maximized by the top-$B$
scores: if a selected candidate has a smaller score than an excluded one,
exchanging them strictly increases the sum. Repeating this exchange produces
Eq.~(21). This conclusion is local and first order; finite-step interactions
between selected candidates are contained in the omitted higher-order term.

\subsection{Policy Instantiations}
\label{app:policy_instantiations}

PAD retains one valuation construction while adapting its supervised interface
to each policy. Let
$m\in\{\mathrm{DD},\mathrm{LTF},\mathrm{MF}\}$ index DiffusionDrive,
LTF, and MeanFuser, respectively, and let
$\operatorname{par}(\cdot)$ denote the parameters of a module. Each
instantiation defines
\begin{equation}
\begin{aligned}
\bar{\ell}_m(z;\beta_m)
&:=\mathbb{E}_{\xi_m}
\left[\ell_m(z;\hat{\theta}_{\mathrm{up},m},
\beta_m,\xi_m)\right],\\
q_i^{(m)}&:=\nabla_{\beta_m}
\bar{\ell}_m(z_i;\hat{\beta}_m).
\end{aligned}
\label{eq:app_policy_interface}
\end{equation}
The same $\bar{\ell}_m$ is used for candidate, calibration, and geometry-set
gradients. This is a deliberate instantiation rather than a general requirement
of influence functions: it prevents auxiliary perception or detection tasks
from changing the meaning of the policy-specific value. Once
Eq.~\eqref{eq:app_policy_interface} is specified, the
derivations in Sections~A.1--A.2 apply without modification.

\begin{table*}[t]
    \centering
    \small
    \setlength{\tabcolsep}{2.5pt}
    \renewcommand{\arraystretch}{1.12}

    \renewcommand{\tabularxcolumn}[1]{m{#1}}

    \begin{tabularx}{\textwidth}{
        >{\bfseries\raggedright\arraybackslash}m{2.35cm}
        >{\raggedright\arraybackslash}X
        >{\raggedright\arraybackslash}X
        >{\centering\arraybackslash}m{1.75cm}
        >{\raggedright\arraybackslash}X
        >{\raggedright\arraybackslash}X
    }
        \toprule

        \textbf{Policy}
        & \makecell[l]{\textbf{Trajectory}\\\textbf{Generation}}
        & \makecell[l]{\textbf{Attribution}\\\textbf{Objective}}
        & \textbf{Randomness}
        & \makecell[l]{\textbf{Attributed}\\\textbf{Parameters}}
        & \makecell[l]{\textbf{Excluded Auxiliary}\\\textbf{Objectives}} \\

        \midrule

        DiffusionDrive
        & Noisy-anchor multimodal refinement
        & $\mathbb{E}_{t,\epsilon}\!\left[
            \mathcal{L}_{\mathrm{focal}}+\mathcal{L}_{1}
          \right]$
        & $t,\epsilon$
        & Trajectory head
        & \makecell[l]{Agent detection;\\BEV semantics} \\

        LTF
        & Direct trajectory regression
        & $w_{\mathrm{traj}}\mathcal{L}_{1}$
        & None
        & Trajectory head
        & \makecell[l]{Agent detection;\\BEV semantics} \\

        MeanFuser
        & MeanFlow proposals followed by ARM reconstruction
        & $\mathbb{E}_{e,r,t}\!\left[
            7\mathcal{L}_{\mathrm{MF}}
            +2\mathcal{L}_{\mathrm{ARM}}
          \right]$
        & $e,r,t$
        & \makecell[l]{MeanFlow and\\ARM heads}
        & BEV semantics \\

        \bottomrule
    \end{tabularx}
    \caption{Policy-specific instantiations of the shared PAD valuation
    interface. Auxiliary losses are excluded only during attribution; final
    curated retraining retains each policy's complete native objective.}
    \label{tab:policy_instantiations}
\end{table*}

\paragraph{DiffusionDrive.}
DiffusionDrive refines a bank of $K$ trajectory anchors. Let $a_k$ be clean
anchor $k$, and let $\mathcal{N}$ and $\mathcal{N}^{-1}$ denote the trajectory
normalization and its inverse. At training time, the head samples a diffusion
timestep $t$ and Gaussian noise $\epsilon_k$, producing
\begin{equation}
\widetilde{a}_{k,t}
=\mathcal{N}^{-1}\!\left(
\operatorname{clip}\!\left[
\sqrt{\bar{\alpha}_t}\,\mathcal{N}(a_k)
+\sqrt{1-\bar{\alpha}_t}\,\epsilon_k,
-1,1\right]\right).
\label{eq:app_dd_noisy_anchor}
\end{equation}
Conditioned on these noisy anchors and the scene representation, decoder stage
$r$ predicts a trajectory $\widehat{\tau}_{rk}$ and a mode logit $c_{rk}$. The
supervision mode is selected from the clean anchors,
\begin{equation}
k^{\star}
=\underset{k\in\{1,\ldots,K\}}{\arg\min}\;
\frac{1}{T}\sum_{s=1}^{T}
\left\lVert\tau^{\star}_{s,xy}-a_{k,s}\right\rVert_2,
\label{eq:app_dd_anchor_assignment}
\end{equation}
and the attribution loss is the native multimodal trajectory objective
\begin{equation}
\begin{aligned}
\ell_{\mathrm{DD}}(z;t,\epsilon)
&=\sum_{r}\left[
w_{\mathrm{cls}}\,
\mathcal{L}_{\mathrm{focal}}(c_r,e_{k^{\star}})\right.\\
&\hspace{3.3em}\left.
+w_{\mathrm{reg}}
\left\lVert\widehat{\tau}_{r k^{\star}}
-\tau^{\star}\right\rVert_1
\right].
\end{aligned}
\label{eq:app_dd_attribution_loss}
\end{equation}
where $e_{k^{\star}}$ is the one-hot mode target; the current configuration
uses $(w_{\mathrm{cls}},w_{\mathrm{reg}})=(10,8)$. Thus,
$\xi_{\mathrm{DD}}=(t,\epsilon)$ and Eq.~(14) of the main paper estimates the
expectation over noisy-anchor realizations. The default attribution subspace is
$\beta_{\mathrm{DD}}=\operatorname{par}(\texttt{\_trajectory\_head})$, which
contains the anchor encoder, timestep conditioning, iterative decoder, and
trajectory prediction layers. Agent-detection and BEV-semantic losses are
excluded because they do not supervise the predicted ego trajectory.

\paragraph{LTF.}
LTF directly regresses one trajectory from its decoded trajectory
query. If $h_{\mathrm{LTF}}(o)$ denotes the frozen upstream scene representation and
$T_{\beta_{\mathrm{LTF}}}$ the trajectory head, then
\begin{equation}
\widehat{\tau}
=T_{\beta_{\mathrm{LTF}}}(h_{\mathrm{LTF}}(o)),
\qquad
\ell_{\mathrm{LTF}}(z)
=w_{\mathrm{traj}}
\left\lVert\widehat{\tau}-\tau^{\star}\right\rVert_1.
\label{eq:app_ltf_attribution_loss}
\end{equation}
The current configuration uses $w_{\mathrm{traj}}=10$. This objective
introduces no policy-specific sampling variable, so
$\xi_{\mathrm{LTF}}=\emptyset$ and its per-example gradient requires no
Monte Carlo averaging. We set
$\beta_{\mathrm{LTF}}=\operatorname{par}(\texttt{\_trajectory\_head})$.
The agent-classification, agent-box, and BEV-semantic terms used during full
training remain outside the attribution objective.

\paragraph{MeanFuser.}
MeanFuser first generates trajectory proposals with MeanFlow and then
reconstructs the final trajectory with the Aggregated Reconstruction Module
(ARM). Write the expert trajectory in the incremental representation
$x=\{(\Delta x_s,\Delta y_s,\sin\theta_s,\cos\theta_s)\}_{s=1}^{T}$. Given
noise $e$ from the trajectory prior and times $0\leq r\leq t\leq1$, MeanFlow
constructs
\begin{equation}
\begin{aligned}
z_t&=(1-t)x+te, \\ 
v_t&=e-x,
\end{aligned}
\label{eq:app_meanflow_target_1}
\end{equation}

\begin{equation}
\begin{aligned}
\widehat{v}_{\beta_{\mathrm{MF}}}
&=u_{\beta_{\mathrm{MF}}}(z_t,r,t;h)\\
&\quad +(t-r)\operatorname{sg}\!\left(
\frac{d}{dt}u_{\beta_{\mathrm{MF}}}(z_t,r,t;h)\right).
\end{aligned}
\label{eq:app_meanflow_target}
\end{equation}
Here, $h$ is the scene context, and the Jacobian--vector correction is stopped
for gradient computation. If $\widehat{x}_{\mathrm{ARM}}$ is the trajectory
reconstructed from the proposal set, the trajectory-specific attribution loss is
\begin{equation}
\begin{aligned}
\mathcal{L}_{\mathrm{MF}}
&=\left\lVert v_t-\widehat{v}_{\beta_{\mathrm{MF}}}\right\rVert_1,\\
\mathcal{L}_{\mathrm{ARM}}
&=\left\lVert\widehat{x}_{\mathrm{ARM}}-x\right\rVert_1,\\
\ell_{\mathrm{MF}}(z;e,r,t)
&=7\mathcal{L}_{\mathrm{MF}}+2\mathcal{L}_{\mathrm{ARM}}.
\end{aligned}
\label{eq:app_meanfuser_attribution_loss}
\end{equation}
Accordingly, $\xi_{\mathrm{MF}}=(e,r,t)$, and independent draws may be averaged
to estimate Eq.~\eqref{eq:app_policy_interface}. The attributed parameters are
\begin{equation}
\begin{aligned}
\beta_{\mathrm{MF}}
&=\operatorname{par}(\texttt{\_meanflow\_head})\\
&\quad\cup\operatorname{par}(\texttt{\_arm\_model\_head}).
\end{aligned}
\label{eq:app_meanfuser_subspace}
\end{equation}
Both modules are required: the first determines the proposal distribution,
whereas the second maps those proposals to the evaluated trajectory. The
auxiliary BEV-semantic loss remains part of full-model training but is excluded
from attribution.

The calibration scenes may be shared across policies, but their outcome
weights cannot. Each $A_j^{(m)}$ is computed from policy $m$'s real-only EPDMS,
and each policy constructs its own gradients, projection, preconditioner, and
score $\psi_i^{(m)}$. PAD therefore compares candidates within a
policy-specific local geometry; absolute values are not directly comparable
across architectures.

\section{Extended Results}
\label{app:extended_results}

\begin{table*}[t]
    \centering
    \small
    \setlength{\tabcolsep}{5.5pt}
    \renewcommand{\arraystretch}{1.08}
    \resizebox{\textwidth}{!}{%
    \begin{tabular}{
        l!{\thinvrule}
        cccc!{\thinvrule}
        ccccc!{\thinvrule}
        cc
    }
            \toprule

            \textbf{Configuration}
            & \textbf{NC}$\uparrow$
            & \textbf{DAC}$\uparrow$
            & \textbf{DDC}$\uparrow$
            & \textbf{TLC}$\uparrow$
            & \textbf{EP}$\uparrow$
            & \textbf{TTC}$\uparrow$
            & \textbf{LK}$\uparrow$
            & \textbf{HC}$\uparrow$
            & \textbf{EC}$\uparrow$
            & \textbf{EPDMS}$\uparrow$
            & \textbf{95\% CI} \\
            \midrule

            \rowcolor{groupgray}
            \textbf{Full pool}
            & 97.65
            & 95.01
            & 99.47
            & 99.78
            & 96.86
            & 96.98
            & 97.08
            & 98.20
            & 69.03
            & 84.31
            & -- \\

            \rowrule

            Remove $G_0$
            & $+0.15$
            & $+0.26$
            & $+0.03$
            & $+0.07$
            & $-0.24$
            & $+0.04$
            & $+0.43$
            & $+0.03$
            & $+4.84$
            & \textbf{$+0.70$}
            & \textbf{$[+0.24,+1.16]$} \\

            Remove $G_1$
            & $+0.78$
            & $+0.72$
            & $+0.09$
            & $+0.02$
            & $-0.82$
            & $+0.58$
            & $-0.12$
            & $+0.12$
            & $+4.47$
            & \textbf{$+1.41$}
            & \textbf{$[+0.97,+1.85]$} \\

            Remove $G_2$
            & $+0.02$
            & $+0.01$
            & $+0.06$
            & $\phantom{+}0.00$
            & $+0.04$
            & $-0.41$
            & $-0.43$
            & $+0.01$
            & $+0.53$
            & $-0.05$
            & $[-0.52,+0.42]$ \\

            Remove $G_3$
            & $+0.42$
            & $+0.17$
            & $-0.30$
            & $+0.05$
            & $-1.03$
            & $+0.68$
            & $+0.33$
            & $+0.12$
            & $-5.72$
            & $-0.40$
            & $[-0.86,+0.07]$ \\

            Remove $G_4$
            & $-0.15$
            & $+0.35$
            & $+0.14$
            & $+0.02$
            & $\phantom{+}0.00$
            & $-0.40$
            & $-0.15$
            & $+0.05$
            & $-14.95$
            & \textbf{$-1.30$}
            & \textbf{$[-1.76,-0.86]$} \\

            \bottomrule
    \end{tabular}%
    }
    \caption{Component-wise NavTest effects of value-ranked group removal. The
    full-pool row reports absolute scores; removal rows report changes relative
    to that reference and paired bootstrap confidence intervals for
    $\Delta$EPDMS.}
    \label{tab:influence_removal_components}
\end{table*}

\begin{table*}[!t]
    \centering
    \setlength{\tabcolsep}{4pt}
    \renewcommand{\arraystretch}{1.08}
    \small
        \begin{tabular*}{\textwidth}{@{\extracolsep{\fill}}
            l!{\thinvrule}
            l!{\thinvrule}
            l!{\thinvrule}
            cccc!{\thinvrule}
            ccccc!{\thinvrule}
            c
        }
            \toprule

            \textbf{Method}
            & \textbf{Data Source}
            & \textbf{Stage}
            & \textbf{NC}$\uparrow$
            & \textbf{DAC}$\uparrow$
            & \textbf{DDC}$\uparrow$
            & \textbf{TLC}$\uparrow$
            & \textbf{EP}$\uparrow$
            & \textbf{TTC}$\uparrow$
            & \textbf{LK}$\uparrow$
            & \textbf{HC}$\uparrow$
            & \textbf{EC}$\uparrow$
            & \textbf{EPDMS}$\uparrow$ \\
            \midrule

            \rowcolor{groupgray}
            \multicolumn{13}{c}{\textit{Diffusion-Based Policy (CVPR 2025)}} \\
            \rowrule

            \multirow{2}{*}{DiffusionDrive}
            & \multirow{2}{*}{Real}
            & S 1
            & 96.6
            & 85.6
            & 99.0
            & 99.3
            & 84.4
            & 95.3
            & 95.1
            & 97.6
            & \textbf{81.8}
            & \multirow{2}{*}{27.9} \\

            &
            & S 2
            & 80.5
            & \textbf{72.2}
            & 84.8
            & 98.7
            & 86.7
            & 76.7
            & 45.9
            & 95.3
            & \textbf{71.8}
            & \\

            \methodrule

            \multirow{2}{*}{SimScale}
            & \multirow{2}{*}{Real+Syn}
            & S 1
            & 96.8
            & 87.1
            & \textbf{99.3}
            & \textbf{99.8}
            & 84.4
            & 96.7
            & \textbf{98.4}
            & 97.6
            & 78.2
            & \multirow{2}{*}{31.3} \\

            &
            & S 2
            & 84.2
            & 70.8
            & 90.8
            & 98.6
            & 91.2
            & 77.3
            & \textbf{61.4}
            & 96.0
            & 39.2
            & \\

            \methodrule

            \multirow{2}{*}{\textbf{PAD}}
            & \multirow{2}{*}{Real+Syn@0.25}
            & S 1
            & 97.8
            & \textbf{89.8}
            & 99.2
            & 99.3
            & \textbf{84.5}
            & 96.4
            & 96.9
            & \textbf{97.8}
            & 75.1
            & \multirow{2}{*}{\epdms{\textbf{32.9}}{+5.0}} \\

            &
            & S 2
            & 83.3
            & 69.7
            & 91.4
            & 98.4
            & 91.0
            & 77.1
            & 59.4
            & 95.1
            & 46.7
            & \\

            \methodrule

            \multirow{2}{*}{\textbf{PAD}}
            & \multirow{2}{*}{Real+Syn@0.50}
            & S 1
            & 96.1
            & 87.3
            & 99.0
            & 99.6
            & 84.3
            & 95.1
            & 97.1
            & \textbf{97.8}
            & 80.0
            & \multirow{2}{*}{\epdms{32.4}{+4.5}} \\

            &
            & S 2
            & \textbf{84.9}
            & 72.0
            & 90.5
            & \textbf{98.9}
            & 90.0
            & \textbf{78.4}
            & 59.3
            & \textbf{96.5}
            & 45.5
            & \\

            \methodrule

            \multirow{2}{*}{\textbf{PAD}}
            & \multirow{2}{*}{Real+Syn@0.75}
            & S 1
            & \textbf{98.3}
            & 89.1
            & 99.0
            & 99.3
            & 84.0
            & \textbf{97.1}
            & 96.7
            & 97.6
            & 74.2
            & \multirow{2}{*}{\epdms{31.7}{+3.8}} \\

            &
            & S 2
            & 83.2
            & 70.9
            & \textbf{91.7}
            & 98.7
            & \textbf{91.3}
            & 76.6
            & 60.8
            & 95.7
            & 42.2
            & \\

            \midrule

            \rowcolor{groupgray}
            \multicolumn{13}{c}{\textit{Regression-Based Policy (NeurIPS 2024)}} \\
            \rowrule

            \multirow{2}{*}{LTF}
            & \multirow{2}{*}{Real}
            & S 1
            & 95.2
            & 78.9
            & 95.1
            & 99.3
            & 81.7
            & 94.4
            & 92.9
            & 97.6
            & 80.0
            & \multirow{2}{*}{23.0} \\

            &
            & S 2
            & 81.8
            & 65.0
            & 83.2
            & 97.7
            & 81.1
            & \textbf{78.9}
            & 47.1
            & \textbf{96.3}
            & \textbf{80.9}
            & \\

            \methodrule

            \multirow{2}{*}{SimScale}
            & \multirow{2}{*}{Real+Syn}
            & S 1
            & \textbf{97.6}
            & \textbf{88.2}
            & \textbf{99.3}
            & \textbf{100.0}
            & 84.1
            & \textbf{96.9}
            & \textbf{96.9}
            & 97.6
            & 78.2
            & \multirow{2}{*}{28.4} \\

            &
            & S 2
            & \textbf{84.6}
            & 64.4
            & \textbf{91.8}
            & 98.0
            & \textbf{93.4}
            & 78.7
            & 59.6
            & 94.7
            & 46.1
            & \\

            \methodrule

            \multirow{2}{*}{\textbf{PAD}}
            & \multirow{2}{*}{Real+Syn@0.25}
            & S 1
            & 97.3
            & 84.2
            & 99.0
            & 99.6
            & \textbf{84.2}
            & 95.8
            & 96.2
            & \textbf{97.8}
            & \textbf{80.9}
            & \multirow{2}{*}{\epdms{28.8}{+5.8}} \\

            &
            & S 2
            & 80.7
            & 69.1
            & 89.8
            & 98.5
            & 89.2
            & 75.9
            & 58.1
            & 95.5
            & 42.7
            & \\

            \methodrule

            \multirow{2}{*}{\textbf{PAD}}
            & \multirow{2}{*}{Real+Syn@0.50}
            & S 1
            & 97.0
            & 82.4
            & 97.7
            & 99.3
            & 83.8
            & 95.8
            & 95.3
            & \textbf{97.8}
            & 76.9
            & \multirow{2}{*}{\epdms{28.8}{+5.8}} \\

            &
            & S 2
            & 82.9
            & 67.6
            & 90.2
            & \textbf{99.0}
            & 90.9
            & 77.5
            & 57.3
            & 95.1
            & 39.4
            & \\

            \methodrule

            \multirow{2}{*}{\textbf{PAD}}
            & \multirow{2}{*}{Real+Syn@0.75}
            & S 1
            & 96.9
            & 85.8
            & 98.9
            & 99.3
            & \textbf{84.2}
            & 95.6
            & 96.7
            & \textbf{97.8}
            & 79.1
            & \multirow{2}{*}{\epdms{\textbf{30.8}}{+7.8}} \\

            &
            & S 2
            & 82.9
            & \textbf{69.3}
            & 90.2
            & 98.7
            & 89.4
            & 78.2
            & \textbf{60.3}
            & 95.2
            & 40.5
            & \\

            \midrule

            \rowcolor{groupgray}
            \multicolumn{13}{c}{\textit{Flow-Matching-Based Policy (CVPR 2026)}} \\
            \rowrule

            \multirow{2}{*}{MeanFuser}
            & \multirow{2}{*}{Real}
            & S 1
            & 96.4
            & 82.0
            & 98.2
            & 99.3
            & 83.3
            & 95.1
            & 94.0
            & \textbf{97.8}
            & 79.1
            & \multirow{2}{*}{24.6} \\

            &
            & S 2
            & 78.0
            & \textbf{71.3}
            & 82.9
            & 98.5
            & 84.7
            & 74.9
            & 47.2
            & 95.7
            & \textbf{76.2}
            & \\

            \methodrule

            \multirow{2}{*}{SimScale}
            & \multirow{2}{*}{Real+Syn}
            & S 1
            & \textbf{97.6}
            & \textbf{86.7}
            & 98.8
            & 99.3
            & 83.9
            & \textbf{96.9}
            & \textbf{96.2}
            & \textbf{97.8}
            & \textbf{80.9}
            & \multirow{2}{*}{\textbf{30.7}} \\

            &
            & S 2
            & 84.5
            & 67.9
            & \textbf{92.8}
            & 98.2
            & \textbf{92.7}
            & \textbf{80.1}
            & \textbf{60.6}
            & 95.4
            & 42.1
            & \\

            \methodrule

            \multirow{2}{*}{\textbf{PAD}}
            & \multirow{2}{*}{Real+Syn@0.25}
            & S 1
            & 96.7
            & 86.4
            & 98.8
            & 99.3
            & 83.3
            & 96.0
            & 94.9
            & 97.6
            & 76.0
            & \multirow{2}{*}{\epdms{30.3}{+5.7}} \\

            &
            & S 2
            & 81.8
            & 70.7
            & 90.6
            & 98.5
            & 89.5
            & 78.0
            & 55.8
            & \textbf{96.8}
            & 52.1
            & \\

            \methodrule

            \multirow{2}{*}{\textbf{PAD}}
            & \multirow{2}{*}{Real+Syn@0.50}
            & S 1
            & \textbf{97.6}
            & 86.0
            & \textbf{99.0}
            & \textbf{99.6}
            & \textbf{84.1}
            & 96.2
            & 94.7
            & 97.6
            & 77.8
            & \multirow{2}{*}{\epdms{30.4}{+5.8}} \\

            &
            & S 2
            & 83.1
            & 69.2
            & 92.1
            & \textbf{98.6}
            & 90.2
            & 78.2
            & 59.4
            & 95.3
            & 47.5
            & \\

            \methodrule

            \multirow{2}{*}{\textbf{PAD}}
            & \multirow{2}{*}{Real+Syn@0.75}
            & S 1
            & 97.4
            & 86.4
            & 98.9
            & 99.3
            & 84.0
            & 96.0
            & 95.8
            & 97.6
            & 80.4
            & \multirow{2}{*}{\epdms{\textbf{30.7}}{+6.1}} \\

            &
            & S 2
            & \textbf{85.0}
            & 67.1
            & 92.5
            & 98.5
            & 89.9
            & 79.0
            & 59.1
            & 95.0
            & 44.2
            & \\

            \bottomrule
        \end{tabular*}%
    \caption{NavHard performance across policy families and selection budgets. PAD recomputes demonstration values for each policy. Stage-specific metrics are reported for both evaluation stages; EPDMS is the combined score.}
    \label{tab:cross_policy_stagewise_results}
\end{table*}

This section complements the aggregate results in the main paper. 
First, group removal evaluates whether PAD values predict the direction of
realized training effects. Second, the complete NavHard results examine whether
policy-specific selection remains effective across policy families, data
budgets, and evaluation stages. These analyses separate valuation fidelity
from downstream effectiveness.

\subsection{Valuation Fidelity}
\label{app:valuation_fidelity}

Figure~2 of the main paper summarizes the relationship between PAD values and realized group-removal effects.
Table~\ref{tab:influence_removal_components} provides the component-level changes and uncertainty estimates. 
We sort the synthetic pool by $\psi$ and partition it into equal-sized groups $G_0,\ldots,G_4$ in ascending order. 
Each run removes one group and retrains DiffusionDrive from the same base checkpoint under an otherwise fixed protocol.
The full-pool run is the common reference, and $\Delta_k$ follows Eq.~(22) of
the main paper. A faithful ranking therefore predicts a negative association:
removing low-valued groups should improve EPDMS, whereas removing high-valued
groups should reduce it.

The observed response is directionally ordered. Nine of the ten pairwise group
comparisons follow the predicted direction, yielding $\rho_s=-0.900$. The
asymptotic Spearman test reported in the main paper gives $p=0.0374$. 
The
prespecified negative alternative gives $p=0.0417$ ($p=0.0833$ two-sided).
Removing both lowest-valued groups improves EPDMS
($\mathrm{Prec@2}=1.00$), whereas removing $G_4$ causes the largest decrease.
This evidence supports directional fidelity, not magnitude calibration:
removing $G_1$ helps more than removing $G_0$, and removing $G_2$ is nearly neutral. 

The component changes explain the aggregate response. Removing $G_0$ or $G_1$
reduces ego progress by $0.24$ or $0.82$ points, respectively, but the gains in
safety, drivable-area compliance, and extended comfort produce positive
$\Delta$EPDMS. Conversely, removing $G_4$ lowers extended comfort by $14.95$
points and also reduces no-collision, time-to-collision, and lane-keeping
scores, yielding $\Delta\mathrm{EPDMS}=-1.30$. PAD values therefore represent
aggregate policy utility under the outcome-conditioned calibration surrogate.

\subsection{Performance Across Policies}
\label{app:cross_policy_results}

Table~\ref{tab:cross_policy_stagewise_results} extends the cross-policy NavTest
evaluation in the main paper to the two-stage NavHard benchmark. PAD recomputes
demonstration values for each policy and compares three selection budgets with
real-only and full-pool training. The analysis tests whether the gains persist
under harder conditions, whether the preferred budget is stable across
benchmarks, and which stage-level effects determine the combined EPDMS.

\paragraph{Aggregate gains.}
Every PAD budget improves upon real-only training. The gains span
$3.8$--$5.0$ EPDMS for DiffusionDrive, $5.8$--$7.8$ for LTF, and
$5.7$--$6.1$ for MeanFuser. At each policy's best budget, PAD reaches $32.9$,
$30.8$, and $30.7$, respectively. Relative to full-pool training, these scores
improve DiffusionDrive by $1.6$ points and LTF by $2.4$ points, while matching
MeanFuser. Policy-aware valuation therefore remains effective across three
training paradigms.

\paragraph{Budget dependence.}
The EPDMS-maximizing budget agrees across NavTest and NavHard: $25\%$ for
DiffusionDrive and $75\%$ for LTF and MeanFuser. DiffusionDrive decreases from
$32.9$ to $32.4$ and $31.7$ as more candidates are retained. LTF instead rises
from $28.8$ to $30.8$ at $75\%$, while MeanFuser increases gradually from
$30.3$ to $30.7$. The opposite responses to additional candidates reinforce
the central premise of PAD: useful data volume is determined jointly by the
candidate pool and the policy that learns from it.

\paragraph{Stage-wise gains and trade-offs.}
The largest improvements over real-only training occur in Stage~2. At each
policy's best budget, driving-direction compliance increases by $6.6$, $7.0$,
and $9.6$ points for DiffusionDrive, LTF, and MeanFuser, respectively;
lane keeping increases by $13.5$, $13.2$, and $11.9$ points; and ego progress
increases by $4.3$, $8.3$, and $5.2$ points. These gains are not uniform:
extended comfort decreases by $25.1$, $40.4$, and $32.0$ points, and
drivable-area compliance also decreases for DiffusionDrive and MeanFuser.
The combined improvements therefore arise from stronger progress and path
adherence rather than component-wise dominance. This result is consistent with
the group-removal analysis: PAD estimates aggregate policy utility and can
retain trade-offs among constituent metrics.

\section{Experimental Details}
\label{app:experimental_details}

This section specifies the controls required to reproduce PAD and its
comparisons. We describe the shared policy interface and policy-specific
valuation settings, followed by calibration, subset construction, full-policy
training, and implementation details. Within each policy family, the
architecture, objective, and optimization schedule are fixed across data
configurations; only the selected synthetic demonstrations vary.

\subsection{Policy Configurations}
\label{app:policy_configurations}

\paragraph{Shared interface.}
All policies receive the same camera-only input: the current left-front,
front, and right-front views are stitched and resized to $2048\times512$, then
combined with the driving command, ego velocity, and ego acceleration. A
ResNet-34 encodes the images under the latent configuration; no LiDAR input is
used. Each policy predicts eight ego-frame poses $(x_s,y_s,\theta_s)$ over
four seconds at $2$ Hz.

\begin{table}[t]
    \centering
    \small
    \setlength{\tabcolsep}{2pt}
    \renewcommand{\arraystretch}{1.08}

    \begin{tabularx}{\columnwidth}{
        >{\raggedright\arraybackslash}X
        >{\centering\arraybackslash}m{1.55cm}
        >{\centering\arraybackslash}m{1.55cm}
        >{\centering\arraybackslash}m{1.55cm}
    }
        \toprule
        \textbf{Setting}
        & \makecell{\textbf{Diffusion-}\\\textbf{Drive}}
        & \textbf{LTF}
        & \textbf{MeanFuser} \\
        \midrule
        \multicolumn{4}{l}{\textbf{Policy configuration}} \\
        Cameras & 3 & 3 & 3 \\
        Resolution & $2048{\times}512$ & $2048{\times}512$ & $2048{\times}512$ \\
        Backbone & R34 & R34 & R34 \\
        Horizon & 4 s & 4 s & 4 s \\
        Frequency & 2 Hz & 2 Hz & 2 Hz \\
        Output poses & 8 & 8 & 8 \\
        Trajectory candidates & 20 & 1 & 8 \\
        Decoder & 2 refinements & Direct & ARM \\
        \midrule
        \multicolumn{4}{l}{\textbf{Valuation configuration}} \\
        Attributed heads & Traj. & Traj. & MF + ARM \\
        Projection dimension $p$ & 2048 & 2048 & 2048 \\
        Geometry-set size $N_h$ & 4000 & 4000 & 2000 \\
        MC samples $K$ & 1 & -- & 1 \\
        Damping $\lambda$ & $10^{-2}$ & $10^{-2}$ & $10^{-2}$ \\
        \bottomrule
    \end{tabularx}
    \caption{Policy and valuation configurations. $N_h$ is the geometry-set
    size, and $K$ is the Monte Carlo sample count per demonstration.}
    \label{tab:policy_valuation_configs}
\end{table}

\paragraph{Policy-specific settings.}
DiffusionDrive refines 20 trajectory anchors through two decoder stages; LTF
directly regresses one trajectory; and MeanFuser reconstructs its output from
eight MeanFlow proposals using ARM. During valuation, gradients are restricted
to the trajectory head for DiffusionDrive and LTF and to the MeanFlow and ARM
heads for MeanFuser. Section~A.3 derives the corresponding objectives, and
Table~\ref{tab:policy_valuation_configs} summarizes the numerical settings.

Within each policy family, all comparisons retain the same architecture,
native objective, and trajectory representation; only the selected synthetic
subset changes. MeanFuser uses a smaller geometry set, while the projection
dimension and damping remain shared across policies.

\subsection{Calibration Protocol}
\label{app:calibration_protocol}

\paragraph{Calibration set.}
We construct a fixed calibration set $\mathcal{C}$ by sampling $M=4000$ real
scenes from a held-out portion of the training split. Synthetic scenes are excluded because their
pseudo-expert quality scores are not equivalent to the downstream outcome of
the evaluated policy. The calibration manifest is fixed before valuation and
shared across policy families.

\paragraph{Policy-specific outcomes.}
For each policy $m$, we evaluate its real-only base checkpoint on every scene
in $\mathcal{C}$ and record the resulting scene-level EPDMS
$s_j^{(m)}$. Outcome weights are computed from that policy's scores using the
global normalization in Eq.~(9) of the main paper. Thus, policies share the
calibration manifest but not EPDMS values or outcome weights. Calibration,
candidate, and geometry-set gradients use the same attribution objective and
parameter subspace defined in Section~A.3.

\paragraph{Separation from selection and evaluation.}
Calibration outcomes and gradients are computed once from the real-only
checkpoint and are not updated during candidate valuation or curated
retraining. The synthetic pool supplies candidates only; NavTest and NavHard
are reserved for final evaluation. Consequently, neither synthetic
pseudo-expert scores nor test outcomes enter the calibration direction,
candidate values, or selected subset.

\begin{table}[t]
    \centering
    \small
    \setlength{\tabcolsep}{2pt}
    \renewcommand{\arraystretch}{1.08}
    \begin{tabularx}{\columnwidth}{
        >{\raggedright\arraybackslash}X
        >{\centering\arraybackslash}m{1.55cm}
        >{\centering\arraybackslash}m{1.55cm}
        >{\centering\arraybackslash}m{1.55cm}
    }
        \toprule
        \textbf{Setting}
        & \makecell{\textbf{Diffusion-}\\\textbf{Drive}}
        & \textbf{LTF}
        & \textbf{MeanFuser} \\
        \midrule
        Optimizer & AdamW & Adam & AdamW \\
        Initial learning rate & $6{\times}10^{-4}$ & $10^{-4}$ & $10^{-4}$ \\
        Schedule & Warmup cosine & Constant & Warmup cosine \\
        Warmup epochs & 3 & -- & 3 \\
        Minimum learning rate & $10^{-6}$ & -- & $10^{-7}$ \\
        Weight decay & $10^{-4}$ & 0 & 0 \\
        Batch size / GPU & 64 & 64 & 64 \\
        Training GPUs & 4 & 4 & 4 \\
        Epochs & 100 & 100 & 100 \\
        \bottomrule
    \end{tabularx}
    \caption{Full-policy training configurations. Batch size is reported per GPU.}
    \label{tab:training_configurations}
\end{table}

\subsection{Training and Baselines}
\label{app:training_baselines}

\paragraph{Baselines.}
At the data level, all variants retain the complete real training set
$\mathcal{D}_{\mathrm{real}}$ and are
distinguished by their use of the synthetic candidate pool. \emph{Vanilla}
uses real data alone. \emph{SimScale (Full)} adds the complete pool, whereas
\emph{SimScale (PDM)} retains only demonstrations generated by the PDM
pseudo-expert. At a synthetic budget $B$, \emph{Random} samples $B$ candidates
uniformly without replacement, while PAD selects the $B$ largest
policy-specific values. Random and PAD therefore use the same source pool and
the same number of synthetic demonstrations at each budget. AutoScale is
included only as a reported external reference because its code and model
weights are unavailable, which is not treated as a controlled baseline.

\paragraph{Full-policy training.}
Selection determines only the synthetic token allowlist. The selected
demonstrations are combined with the real set, and all policy parameters are
optimized with the policy's native supervised objective. The parameter
restriction used during valuation therefore does not constrain final training.
Table~\ref{tab:training_configurations} summarizes the locally trained
configurations. Each run uses cached features and distributed data parallelism
on four A800 GPUs for 100 epochs.

\paragraph{Evaluation.}
Trained policies are evaluated using the official NAVSIM
one-stage protocol on NavTest and two-stage protocol on NavHard. We use four
L40 GPUs and report EPDMS on a $0$--$100$ scale; all component metrics in the
stage-wise tables use the same scaling. Neither evaluation split contributes
to training, calibration, or candidate ranking.

\subsection{Reproducibility}
\label{app:reproducibility}

\paragraph{Randomness and data identity.}
Policy training uses seed 0 through the PyTorch Lightning seeding interface,
including data-loader workers. Calibration-set construction and the random
baseline use seed 2026, while geometry-set sampling and Gaussian
projection use seed 0. We persist the calibration manifest, policy-specific
value file, and selected-token allowlist. These artifacts identify every scene
used for calibration and final retraining, preventing cache-layout changes
from altering the selected data.

\paragraph{Per-example gradients.}
Valuation features are computed from cached samples with an effective batch
size of one. Unused parameters receive zero gradients before flattening, and
the geometry set, calibration set, and candidates use the same
checkpoint, parameter ordering, and projection within a policy. DiffusionDrive
and MeanFuser remain in training mode because their attribution objectives use
the training path; their BatchNorm layers are separately fixed in evaluation
mode to preserve stored running statistics. LTF is evaluated entirely in
evaluation mode. We use one stochastic loss realization for DiffusionDrive and
MeanFuser, as reported in Table~\ref{tab:policy_valuation_configs}.

\paragraph{Projection and numerical solution.}
The Gaussian projection is regenerated deterministically from seed 0 in
chunks of 262,144 parameters. This avoids materializing the full projection
matrix while preserving the same map for geometry-set, calibration, and
candidate gradients. Samples are loaded sequentially from cache, and only their projected
rows are retained. After constructing the damped geometry with the settings in
Table~\ref{tab:policy_valuation_configs}, we aggregate the policy-aware
calibration direction and solve the projected linear system associated with
Eq.~(18) of the main paper using \texttt{torch.linalg.solve}. A single
matrix--vector product then
produces all candidate values, which are stored with token and source metadata.

\section{Qualitative Results}
\label{app:qualitative}

The aggregate results are complemented with scene-level trajectory
visualizations. Each example combines the bird's-eye-view context, the three camera views, predicted trajectories, and the relevant scene-level scores.

\subsection{Planning Comparisons}
\label{app:planning_comparisons}

\begin{figure*}[t]
    \centering
    \includegraphics[width=\linewidth]{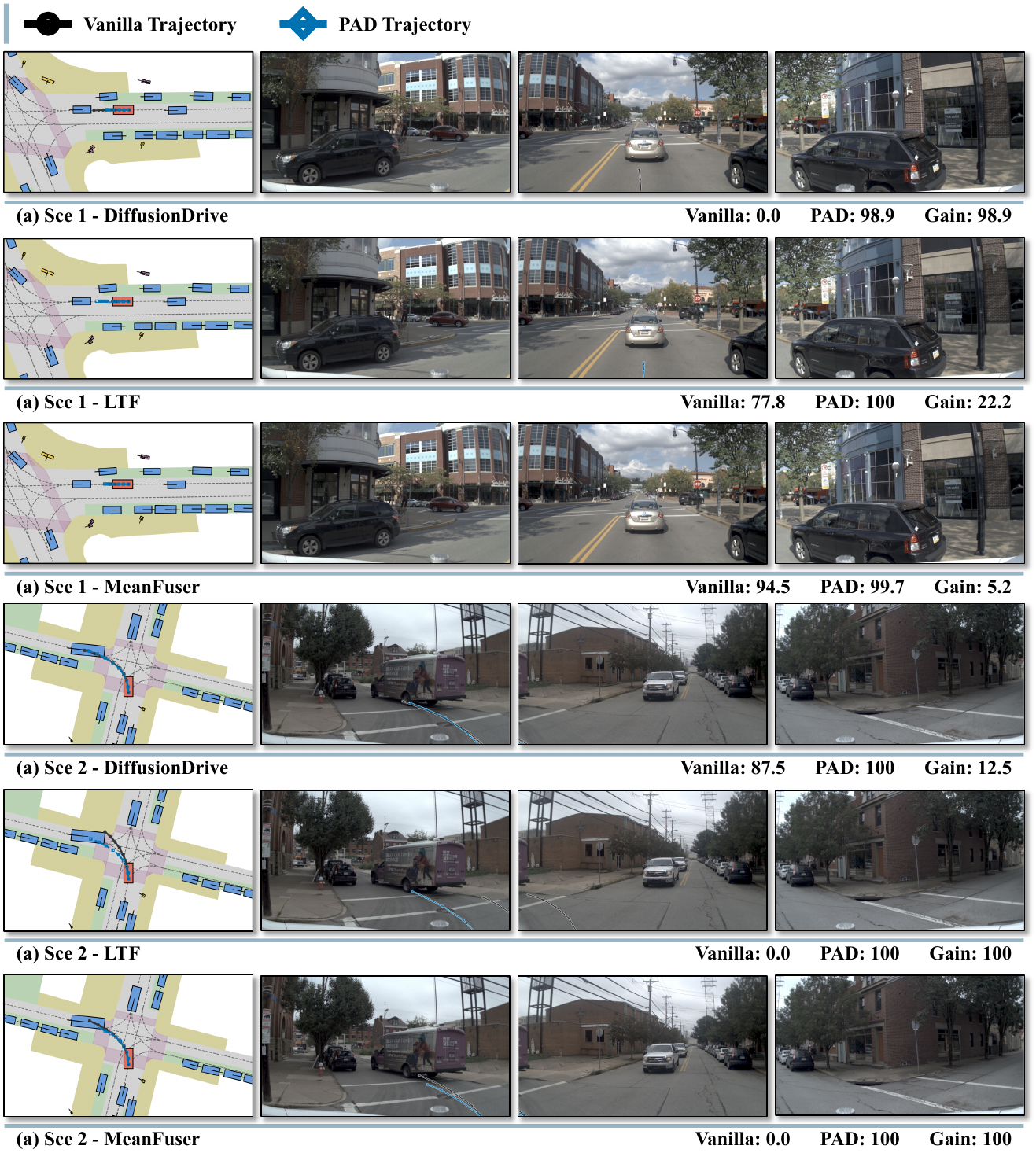}
    \caption{Cross-policy improvements on two representative NavTest scenes.
Within each scene, rows show DiffusionDrive, LTF, and MeanFuser under real-only training (Vanilla, black) and PAD curation (blue). }
    \label{fig:qualitative_cross_policy}
\end{figure*}

\paragraph{Cross-policy improvements.}
Figure~\ref{fig:qualitative_cross_policy} applies DiffusionDrive, LTF, and MeanFuser to the same scenes. 
PAD improves every policy in both scenes, despite their different trajectory parameterizations and objectives. 
In the first scene, EPDMS increases from $0.0$ to $98.9$ for DiffusionDrive, from $77.8$ to $100$ for LTF, and from $94.5$ to $99.7$ for MeanFuser. 
In the second, PAD raises DiffusionDrive from $87.5$ to $100$ and resolves zero-score predictions from both LTF and MeanFuser. 
These shared improvements show that policy-specific valuation can identify training data that benefits distinct policy families without requiring a common output representation.

\begin{figure*}[t]
    \centering
    \includegraphics[width=\linewidth]{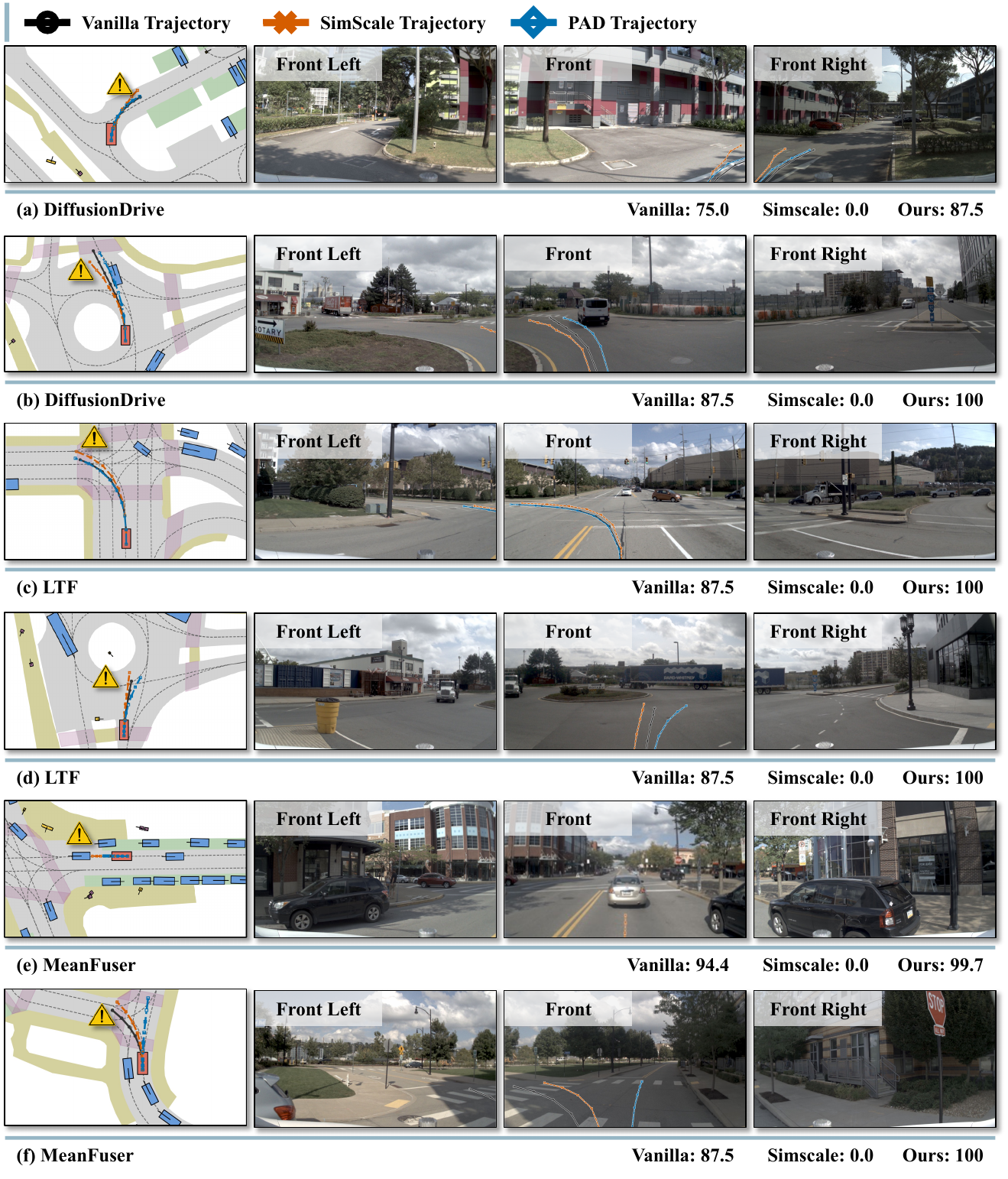}
    \caption{Representative failures induced by full-pool synthetic training.
Rows show two scenes per policy. Black, orange, and blue denote the real-only, full-pool SimScale, and PAD trajectories, respectively.}
    \label{fig:qualitative_full_pool}
\end{figure*}

\paragraph{Protection against harmful scaling.}
Figure~\ref{fig:qualitative_full_pool} compares real-only training, full-pool SimScale, and PAD. 
Full-pool augmentation produces a zero EPDMS in all six
examples, although the corresponding real-only policies remain partially or fully viable. 
PAD avoids these failures and also exceeds the real-only score, reaching $87.5$--$100$ across the three policy families. 
The contrast isolates the role of composition that indiscriminate synthetic augmentation can move a policy away from a feasible solution, whereas valuation-based curation preserves useful supervision while excluding detrimental examples.

\begin{figure*}[t]
    \centering
    \includegraphics[width=\linewidth]{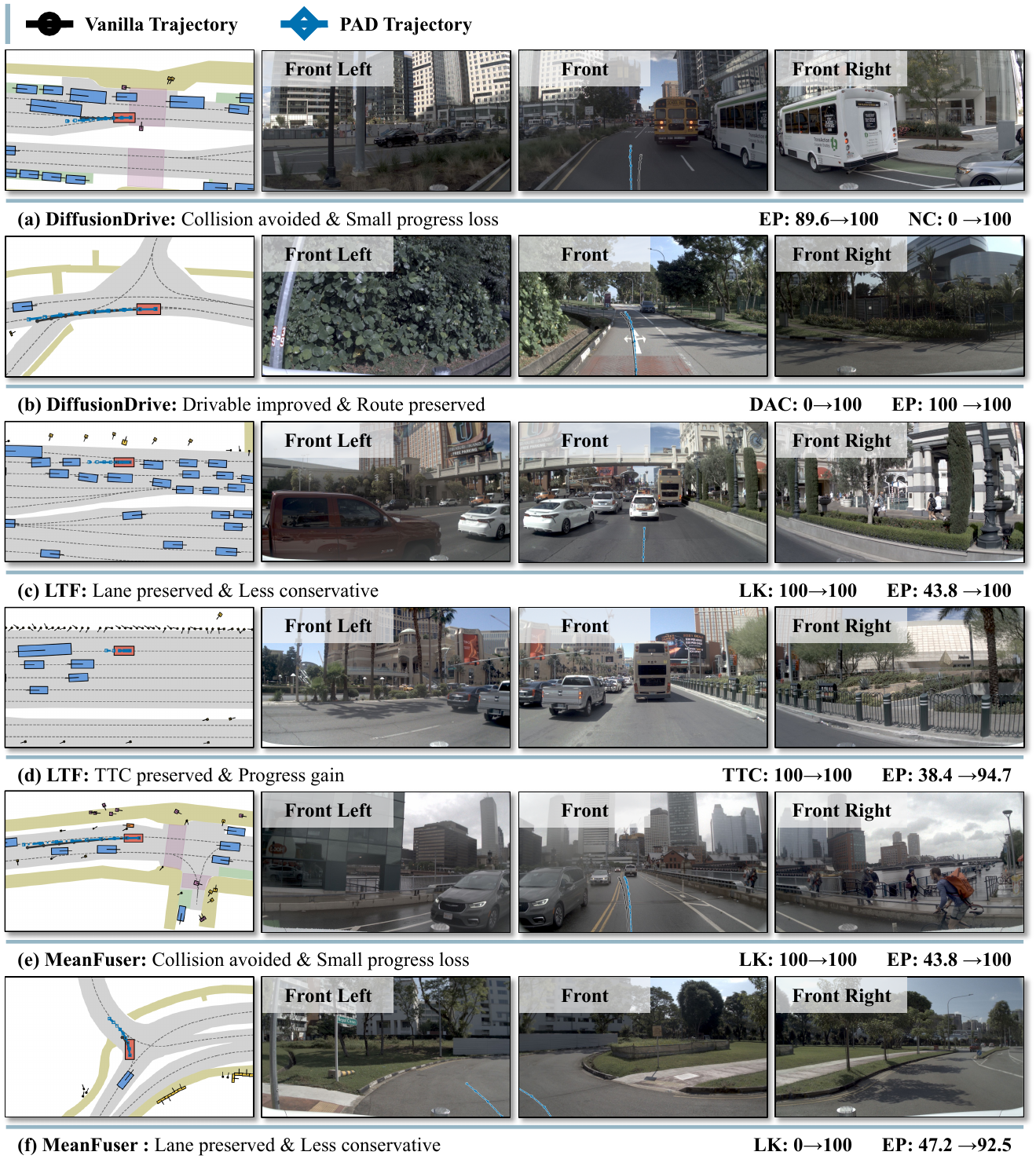}
    \caption{Component-wise effects of PAD. Black and blue denote the real-only and PAD trajectories. Metric arrows report the change from real-only training to PAD. 
Across policies, PAD improves safety, drivable-area compliance, lane keeping, or progress without degrading the paired component shown in each row.}
    \label{fig:qualitative_components}
\end{figure*}

\paragraph{Component-wise balance.}
The examples in Figure~\ref{fig:qualitative_components} show that the aggregate gain does not arise from exchanging one desirable property for another. 
For DiffusionDrive, PAD raises no-collision or drivable-area compliance from $0$ to $100$ while preserving or improving ego progress. 
For LTF, ego progress increases from $43.8$ to $100$ and from $38.4$ to $94.7$, while lane keeping and time-to-collision compliance remain at $100$. 
For MeanFuser, PAD either preserves lane keeping while improving progress or improves both components simultaneously. 

\subsection{Failure Cases}
\label{app:failure_cases}

\begin{figure*}[t]
    \centering
    \includegraphics[width=\linewidth]{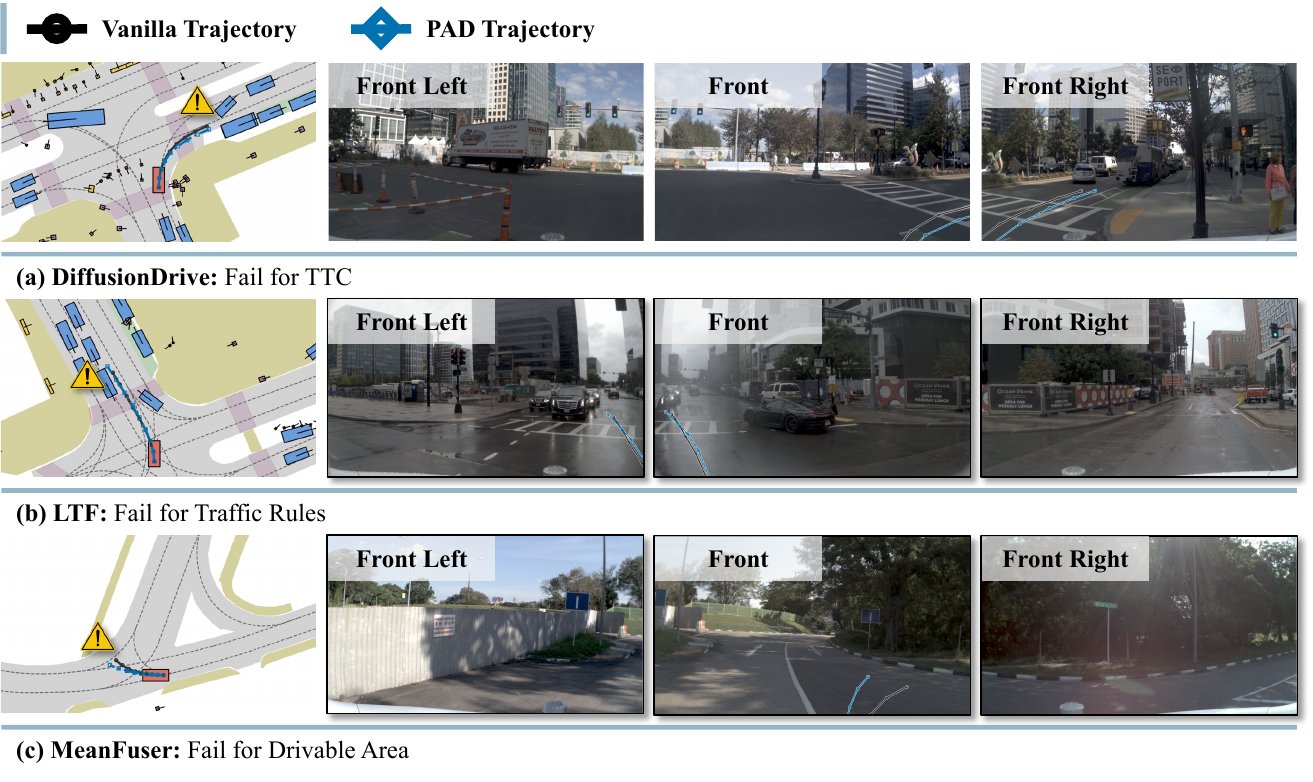}
    \caption{Representative unresolved failures for DiffusionDrive, LTF, and
    MeanFuser. Black and blue denote the real-only and PAD trajectories. Both
    variants fail the indicated time-to-collision, traffic-rule, or
    drivable-area criterion.}
    \label{fig:qualitative_failures}
\end{figure*}

PAD improves data composition but cannot eliminate errors caused by unresolved policy or scene complexity. Figure~\ref{fig:qualitative_failures} presents one such case for each policy family. Neither the real-only nor PAD-trained DiffusionDrive satisfies the time-to-collision criterion; LTF continues to violate traffic rules in a complex intersection; and MeanFuser leaves the drivable area near an irregular junction. 
PAD changes the predicted trajectory in each case but does not fully restore feasibility. 

\end{document}